\documentclass[letterpaper]{article} 
\usepackage{aaai24}  
\usepackage{times}  
\usepackage{helvet}  
\usepackage{courier}  
\usepackage[hyphens]{url}  
\usepackage{graphicx} 
\usepackage{natbib}  
\usepackage{caption} 
\usepackage{algorithm}
\usepackage{algorithmic}

\usepackage{dsfont}
\usepackage{subfigure}
\usepackage{booktabs}
\usepackage{multirow}
\usepackage{color}
\usepackage{makecell}
\usepackage{tabularx}
\usepackage{listings}
\usepackage{enumerate}
\usepackage{amsmath}
\usepackage{bbding}
\usepackage{pifont}
\usepackage{wasysym}
\usepackage{amssymb}

\usepackage{soul}
\usepackage{color, xcolor} 
\usepackage{paralist}

\usepackage{newfloat}
\usepackage{listings}
\DeclareCaptionStyle{ruled}{labelfont=normalfont,labelsep=colon,strut=off} 
\floatstyle{ruled}
\newfloat{listing}{tb}{lst}{}
\floatname{listing}{Listing}
\title{Get an A in Math: Progressive Rectification Prompting}
\author {
    Zhenyu Wu\textsuperscript{\rm 1},
    Meng Jiang\textsuperscript{\rm 2},
    Chao Shen\textsuperscript{\rm 1}
}
\affiliations {
    \textsuperscript{\rm 1}School of Cyber Science and Engineering, Xi’an Jiaotong University\\
    \textsuperscript{\rm 2}Department of Computer Science and Engineering, University of Notre Dame, Notre Dame, IN 46556\\
    zhenyuwu@stu.xjtu.edu.cn, mjiang2@nd.edu, chaoshen@xjtu.edu.cn
}

\usepackage{bibentry}

\begin{document}

\maketitle

\begin{abstract}
Chain-of-Thought (CoT) prompting methods have enabled large language models (LLMs) to generate reasoning paths and solve math word problems (MWPs).
However, they are sensitive to mistakes in the paths, as any mistake can result in an incorrect answer.
We propose a novel method named Progressive Rectification Prompting (PRP) to improve average accuracy on eight MWP datasets from $77.3$ to $90.5$. Given an initial answer from CoT, PRP iterates a verify-then-rectify process to progressively identify incorrect answers and rectify the reasoning paths.
With the most likely correct answer, the LLM predicts a masked numerical value in the question; if the prediction does not match the masked value, the
answer is likely incorrect. Then the LLM is prompted to re-generate the reasoning path hinted with a set of incorrect answers to prevent itself from repeating previous mistakes. PRP achieves the best performance compared against the CoT methods.
Our implementation is made publicly available at \url{https://wzy6642.github.io/prp.github.io/}.
\end{abstract}

\vspace{-0.1in}
\section{Introduction}
\label{sec:intro}
Math word problems (MWPs) require language comprehension, mathematical reasoning, and problem-solving skills. Studying these problems helps AI researchers develop algorithms and models that can mimic human-like reasoning and problem-solving abilities \cite{chen-2022-pot}.
Chain-of-thought (CoT) prompting methods help large language models (LLMs) break down complex problems into manageable parts, allowing them to focus on each part individually \cite{Kojima-2022-CoT}. The LLMs become decent zero-shot reasoners by simply adding ``\emph{Let's think step by step}'' to generate reasoning paths and predict answers to the MWPs \cite{freda-2023-gsm8k-ic,wang-2023-ps,wang-2023-selfconsistency,zheng-2023-php}.

When analyzing the performance of existing methods, we found that the average accuracy on eight standard datasets (e.g., MultiArith, GSM8K) was $77.3$, far below A-level grades. Because they have three drawbacks: (1) lack of verification that checks if the answer is correct, (2) lack of rectification that finds the correct answer being aware of mistakes, and (3) lack of an effective method that progressively refines reasoning path, which are essential ``exam skills.'' 

First, to distinguish correct and incorrect answers, existing methods repeatedly solve a problem and use a majority vote strategy to determine the most consistent answer as the correct answer. This is known as self-consistency \cite{wang-2023-selfconsistency}.
However, since it solves the same problem multiple times, this repeated independent process leads to same mistakes, making the frequent answer still incorrect.
Second, existing methods such as progressive-hint prompting \cite{zheng-2023-php} modify reasoning paths by adding ``\emph{(Hint: The answer is near \emph{$\mathsf{[H]}$})}'' after the given problem, where $\mathsf{[H]}$ is the slot of previous answers. It is evident that when previous answers are incorrect, LLMs may still generate an incorrect answer in response to the hint.
Third, existing CoT prompting methods exhibit high sensitivity to mistakes in intermediate reasoning steps \cite{Kojima-2022-CoT,chen-2022-pot,wang-2023-ps,freda-2023-gsm8k-ic}. Even a tiny mistake in the reasoning process could alter the entire problem-solving process, resulting in an incorrect answer. It is nontrivial to achieve multi-step precise reasoning.

\begin{figure*}[t]
  \centering
  \includegraphics[width=0.99\textwidth]{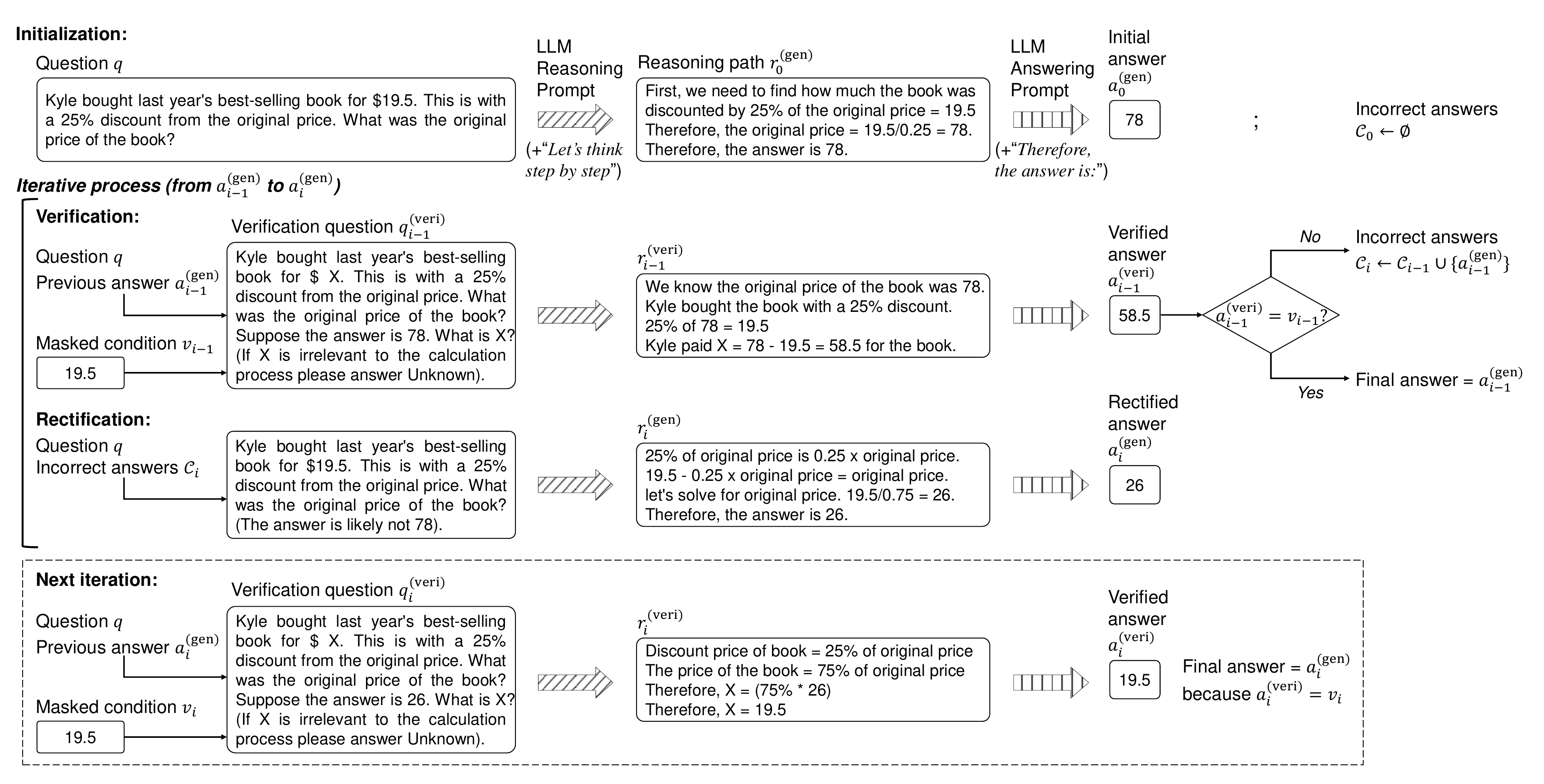}
  \caption{Overview of Progressive Rectification Prompting (PRP) method. PRP first generates an initial answer. PRP then iterates a verify-then-rectify process to progressively rectify the LLM-generated answer to find the correct one.}
  \label{fig:framework}
  \vspace{-0.13in}
\end{figure*}

To address the three drawbacks of existing methods, our research is inspired by a guide on math study skills and exam success written by \citeauthor{gall1990tools} in 1990.
First, \emph{substitute verification} has been commonly used in math exams to verify the correctness of an answer. Let us look at a specific example. Given an equation $2 + 3 = y$, after solving it, we find that the answer $y$ equals $5$. Next, we introduce a masked condition to formulate the masked equation $X + 3 = y$ and substitute the answer $5$ into this equation. Solving the masked equation, if $X$ equals $2$, it indicates that the answer $5$ align with the original equation. So the answer was more likely to be correct. Otherwise, the answer could probably be incorrect. Compared to repeatedly checking, such as solving the same question multiple times, the substitute verification can effectively prevent the repetition of mistakes and improve the accuracy of answer verification.
Second, relying on existing progressive hints such as ``\emph{the answer is near \emph{$\mathsf{[H]}$}}'' would limit the exploration of other potential answers when the hint answer was incorrect. Suppose that $\mathsf{[H]}$ has been found less likely correct by substitute verification. A negation hint on it, like ``\emph{the answer is likely \textbf{not} \emph{$\mathsf{[H]}$}}'', will help LLMs eliminate or less consider such answers, so the LLMs will actively rectify their reasoning paths to avoid mistakes.
Third, the \emph{dual process theory} in psychology \cite{Jonathan-2003-two-stage} tells us that humans have two cognitive systems to progressively refine their answers, plans, and solutions. One provides initial responses based on intuition; the other provides a deliberate and reflective approach to progressive refine those initial responses.
Existing CoT prompting methods possess only the capability of the first system, while lacking the capacity for progressive refinement of answers through the second type.

In this paper, we propose a novel zero-shot prompting method to implement and integrate the above ideas to improve the performance of LLMs on MWPs.
We name it Progressive Rectification Prompting (PRP).
Figure~\ref{fig:framework} illustrates PRP with an example from the GSK8K dataset.
In PRP, an initial answer is generated by a standard zero-shot prompt \cite{Kojima-2022-CoT}.
Then PRP feeds the question and initial answer into an iterative \textit{verify-then-rectify} process.
It progressively rectifies the LLM-generated answer to find the correct one.
The verify-then-rectify process consists of a \emph{verification} module and a \emph{rectification} module.
The verification module uses the substitute verification method to verify the correctness of the answer. It masks a numerical value in the question, takes the previous generated answer as a conclusion, and uses it as a new condition. If the masked value is predicted incorrectly, the answer is added to the set of potentially incorrect answers.
The rectification module designs a hint that uses the set of potentially incorrect answers as feedback to rectify previous answers.
In Figure~\ref{fig:framework}, the initial numerical answer was $78$.
Next, PRP used a regular expression to match all numerical values within the question. Then it randomly selected one of these values (i.e., $19.5$ in this example) and replaced its occurrence in the question with a special token X. This converted the known condition in the original question into an unknown condition, resulting in the masked question. Subsequently, we rewrote the masked question using a simple template to form the verification question: ``$\mathsf{[Q]}$ Suppose the answer is $\mathsf{[A]}$. What is $\textmd{X}$? (If $\textmd{X}$ is irrelevant to the calculation process please answer Unknown)'', where $\mathsf{[Q]}$ was the slot for the masked question, and $\mathsf{[A]}$ was the slot for previous generated answer. If the answer did not match the masked condition, the previous generated answer would be considered less likely correct and added to a set of potentially incorrect answers.
In rectification, we added the phrase ``(The answer is likely not $\mathsf{[H]}$)'' after the given question, where $\mathsf{[H]}$ was the slot for the set of potentially incorrect answers. The LLM avoided repeating previous mistakes when re-answering the question using the set of potentially incorrect answers as feedback.
In most cases, the LLM got the correct answer with a single rectification. But to deal with complex arithmetic questions, PRP had to iterate the verify-then-rectify process to progressively rectify the answer.

Experiments on text-davinci-003 demonstrate that the proposed PRP method improves over existing prompting methods by a striking margin across eight MWP datasets. Our method attains an average score of $90.5$, significantly higher than $77.3$ from the best of zero-shot CoT, and even higher than $81.0$ from the best of few-shot CoT. Our PRP equips LLMs with high-level math exam skills.

The main contributions are summarized as follows: 
\begin{compactitem}
\item We propose a novel zero-shot prompting method that enables LLMs to progressively rectify the generated answer and accurately solve math word problems. It has an iterative verify-then-rectify process to avoid repeating previous mistakes and achieve continuous improvement.
\item We conduct extensive experiments on eight math word problem datasets under zero-shot and few-shot CoT settings. Notably, our method achieves the state-of-the-art performance and attains an A-level grade on average.
\end{compactitem}

\section{Related Work}
\label{sec:related}
\subsection{Math Word Problem Solving}
Our work is related to existing efforts in solving math word problems (MWPs).
Traditional methods used statistical learning-based approaches to extract entities, quantities, and operators from a question and generated an arithmetic equation to find the answer \cite{Hosseini-2014-AddSub, roy-etal-2015-singleop, zhou-2015-qp, mitra-2016-mathstudent}.
Recent methods based on sequence-to-sequence (Seq2Seq) model and recurrent neural networks directly transformed the question into an arithmetic equation \cite{wang-2017-dns, wang-2019-template} . 
However, their generated equations could be invalid or unsolvable.
Besides, recent efforts fine-tuned pre-trained language models on a variety of downstream tasks \cite{shen-2021-generate-rank, liang-2022-mwp, liang-2023-wda}, which significantly improved the validity of generated equations and brought substantial performance improvements over Seq2Seq models.
These methods require a significant amount of human annotations, lacking the ability to generalize to new MWP datasets.
In this work, we aim to directly prompt the LLMs to answer arbitrary MWPs without human annotation and task-specific fine-tuning.
Our method can generate reasoning paths that enable researchers to investigate model behavior.

\subsection{Chain-of-Thought Prompting Methods}
Our work is related to Chain-of-Thought (CoT) prompting methods, which enable LLMs to generate reasoning paths and solve MWPs.
Two types of CoT prompting methods have been proposed: zero-shot prompting \cite{Kojima-2022-CoT} and few-shot prompting.
By adding ``\emph{Let's think step by step}'' after the question and feeding the modified question to the LLMs, the LLMs can generate complex reasoning paths.
However, zero-shot CoT prompting suffers from missing-step errors. To mitigate these errors, Plan-and-Solve (PS) prompting method instructed the LLMs to devise a plan for breaking down the entire task into smaller subtasks, and then carry out the subtasks according to the plan \cite{wang-2023-ps}.
All these methods are based on manually writing instructions, to eliminate human labor, \citeauthor{zhang-2023-auto-instruct} proposed Auto-Instruct to automatically improve the quality of instructions provided to LLMs.
Manual-CoT \cite{wei-2022-cot}, as a type of few-shot promopting, designed effective manual demonstrations to elicit multi-step reasoning ability of LLMs.
Program of Thought (PoT) \cite{chen-2022-pot} used LLMs to generate programming language statements, and used a program interpreter to execute the generated program to get the final answer.
To leverage the benefit of demonstration examples and minimize manual effort, \citeauthor{zhang-2023-autocot} designed Auto-CoT. By sampling questions with diversity and generating reasoning path to automatically construct demonstrations. 
\citeauthor{yu-2023-ifqa} introduced IfQA, a dataset for counterfactual reasoning, which requires models to identify the right information for retrieval and inference.
These methods are sensitive to mistakes in reasoning paths, and any mistake can result in an incorrect answer.
Our method iterates a verify-then-rectify process to progressively identify incorrect answers and rectify reasoning paths.

\subsection{Answer Selection}
Several studies have trained models to evaluate candidate answers and select the best answer as the final response.
For example, 
\citeauthor{kushman-2014-math} trained a classifier to select the best answer from candidate answers. 
\citeauthor{roy-roth-2015-multiarith} trained a relevance classifier and a lowest common ancestor operation classifier.
The distributional output of these classifiers was used in a joint inference procedure to determine the final answer.
\citeauthor{shen-2021-generate-rank} jointly trained a candidate expression generator and a candidate expression ranker to get better answers.
\citeauthor{karl-2021-gsm8k} fine-tuned GPT-3 as a scorer to calculate solution-level verification score and choose the highest score answer as the final answer.
All these methods require massive human annotations.
In contrast, our method automatically verifies the correctness of LLM-generate answers and selects the answer that has been verified.

\section{Proposed Method}
\label{sec:method}

\begin{table}[t]
\centering
\renewcommand\arraystretch{1.1}
  \caption{Notations and their definitions.}
  \resizebox{1.0\columnwidth}{!}{
  \begin{tabular}{l|l}
    \bottomrule
    {\bf Notation} & {\bf Definition} \\
    \hline
    $q$ & Math word problem \\
    $q_i^{(\text{veri})}$ & Verification question  \\
    $r_i^{(\text{gen})}$ & LLM-generated reasoning path for question $q$ \\
    $a_i^{(\text{gen})}$ & LLM-generated answer for question $q$ \\
    $r_i^{(\text{veri})}$ & LLM-generated reasoning path for question $q_i^{(\text{veri})}$ \\
    $a_i^{(\text{veri})}$ & LLM-generated answer for question $q_i^{(\text{veri})}$ \\
    $v_i$ & Masked condition \\
    $\mathcal{C}_i$ & The set of potentially incorrect answers \\
    $\mathsf{[Q]}$ & Slot for question \\
    $\mathsf{[R]}$ & Slot for reasoning path \\
    $\mathsf{[A]}$ & Slot for answer \\
    $\mathsf{[H]}$ & Slot for set of potentially incorrect answers \\
    \toprule
\end{tabular}  }
\vspace{-0.2in}
\label{tab:notation}
\end{table}

\subsection{Overview}
We propose a novel zero-shot prompting method named Progressive Rectification Prompting (PRP) for solving math word problems.
Figure~\ref{fig:framework} illustrates the PRP method.
Given a question $q$, PRP prompts the LLM to generate the final answer. 
Specifically, it first prompts the LLM to generate an initial answer $a_0^{(\text{gen})}$ and initializes the set of potentially incorrect answers as an empty set $\mathcal{C}_0=\varnothing$.
Then, it iterates the verify-then-rectify process up to $K$ iterations to progressively rectify the LLM-generated answer.
This process consists of a verification module and a rectification module.
In the $i$-th iteration, the verification module uses the substitute verification method to verify the correctness of the previous generated answer $a_{i-1}^{(\text{gen})}$.
If the answer $a_{i-1}^{(\text{gen})}$ is verified likely to be incorrect, 
add $a_{i-1}^{(\text{gen})}$ to the set of potentially incorrect answers $\mathcal{C}_{i-1}$ to obtain the updated set $\mathcal{C}_{i}$.
Otherwise, take $a_{i-1}^{(\text{gen})}$ as the final answer.
The rectification module uses the set of potentially incorrect answers $\mathcal{C}_{i}$ as feedback to rectify previous answers and generate the rectified answer $a_{i}^{(\text{gen})}$.
If the number of iterations exceeds the maximum iteration $K$, take the last LLM-generated answer $a_{K}^{(\text{gen})}$ as the final answer. 
In the following sections, we will elaborate on the details of each component.
Table \ref{tab:notation} presents a list of the notations used throughout this paper.

\subsection{Initialization} 
During initialization, PRP initializes the set of potentially incorrect answers as an empty set $\mathcal{C}_0=\varnothing$ and prompts the LLM to generate an initial answer $a_0^{(\text{gen})}$ for the given question $q$.
Specifically, we first construct a reasoning generation prompt: ``Q: $\mathsf{[Q]}$. A: Let's think step by step'', 
where $\mathsf{[Q]}$ is the slot for question $q$. 
We then feed the above prompt to the LLM, which subsequently generates a reasoning path $r_0^{(\text{gen})}$.

To extract the answer from the reasoning path, we add the answer extraction instruction after the reasoning path to devise the answer generation prompt: ``$\mathsf{[R]}$ Therefore, the answer (expressed in Arabic numerals and without units) is:'', 
where $\mathsf{[R]}$ is the slot for reasoning path $r_0^{(\text{gen})}$. 
Finally, we feed the answer extraction prompt to the LLM to generate the initial answer $a_0^{(\text{gen})}$ for the question $q$.

\subsection{Iterative Verify-then-Rectify Process}
We propose a novel iterative verify-then-rectify method that progressively rectifies the LLM-generated answer over $K$ iterations by cyclic execution of the verification and rectification modules.
The iteration process would terminate early if the LLM-generated answer is verified likely to be correct. 
Here we take the $i$-th iteration as an example to illustrate the verify-then-rectify process.

\subsubsection{Verification Module}
The verification module uses substitute verification method to verify the correctness of the previous generated answer $a_{i-1}^{(\text{gen})}$. 
It comprises several substeps.

Firstly, we utilize the condition mask method \cite{weng-2022-verification} to create a masked question. 
Specifically, we first use a regular expression to match all numerical values within the question $q$.
We then randomly select one of these values $v_{i-1}$ and replace its occurrence in the question $q$ with a special token X, resulting in the masked question. 

Secondly, we rewrite the masked question using a simple template to form the verification question $q_{i-1}^{(\text{veri})}$:  ``$\mathsf{[Q]}$ Suppose the answer is $\mathsf{[A]}$, what is X? (If X is irrelevant to the calculation process please answer Unknown)'', 
where $\mathsf{[Q]}$ is the slot for masked question, and $\mathsf{[A]}$ is the slot for previous generated answer $a_{i-1}^{(\text{gen})}$. 

Thirdly, we feed the reasoning generation prompt ``Q: $\mathsf{[Q]}$. A: Let's think step by step'' into the LLM to generate a reasoning path $r_{i-1}^{(\text{veri})}$ for the verification question $q_{i-1}^{(\text{veri})}$, 
where $\mathsf{[Q]}$ is the slot for question $q_{i-1}^{(\text{veri})}$.
Furthermore, we feed the answer generation prompt ``$\mathsf{[R]}$ Therefore, the answer (expressed in Arabic numerals and without units) is:'' into the LLM to generate the answer $a_{i-1}^{(\text{veri})}$ for the verification question $q_{i-1}^{(\text{veri})}$.
Where $\mathsf{[R]}$ is the slot for reasoning path $r_{i-1}^{(\text{veri})}$.

Finally, we check if $a_{i-1}^{(\text{veri})}$ is equal to $v_{i-1}$. 
If they are equal, it indicates that the previous generated answer $a_{i-1}^{(\text{gen})}$ is most likely correct.
We select $a_{i-1}^{(\text{gen})}$ as the final answer and exit the loop.
Otherwise, the previous generated answer $a_{i-1}^{(\text{gen})}$ is likely incorrect, and we add $a_{i-1}^{(\text{gen})}$ to the set of potentially incorrect answers $\mathcal{C}_{i-1}$ to obtain the updated set $\mathcal{C}_{i}$.


\subsubsection{Rectification Module}
The rectification module uses a set of potentially incorrect answers $\mathcal{C}_i=\{a_{0}^{(\text{gen})}, \cdots, a_{i-1}^{(\text{gen})}\}$ as feedback to generate a rectified answer $a_{i}^{(\text{gen})}$.
Specifically, we first devise an answer rectification prompt: ``Q: $\mathsf{[Q]}$ (The answer is likely not $\mathsf{[H]}$) A: Let's think step by step'', 
where $\mathsf{[Q]}$ is the slot for the question $q$, and $\mathsf{[H]}$ is the slot for the set of potentially incorrect answers $\mathcal{C}_i$. 
We then feed the above prompt into the LLM to generate a rectified reasoning path $r_{i}^{(\text{gen})}$. 
Finally, we feed the prompt ``$\mathsf{[R]}$ Therefore, the answer (expressed in Arabic numerals and without units) is:'' into the LLM to generate the rectified answer $a_{i}^{(\text{gen})}$ for the question $q$.
Where $\mathsf{[R]}$ is the slot for reasoning path $r_{i}^{(\text{gen})}$.

\subsubsection{Answer Selection}
The process of verify-then-rectify can be iterated until specific stopping conditions are met.
The process terminates under two situations.
The first is when the answer $a_{i-1}^{(\text{gen})}$ is verified likely to be correct.
In this case, we select answer $a_{i-1}^{(\text{gen})}$ as the final answer. 
The second situation is when the number of iterations exceeds the maximum iteration $K$. 
In this case, we choose the last LLM-generated answer $a_{K}^{(\text{gen})}$ as the final answer.

\section{Experiments}
\label{sec:experiments}
\begin{table*}[t]
\renewcommand\arraystretch{1.35}
\centering
\caption{Accuracy comparison on eight math word problem datasets. The best and second best results are boldfaced and underlined,  respectively. All indicators are presented in percentages.}
\resizebox{\textwidth}{!}{
\begin{tabular}{l|l|p{1.7cm}<{\centering}p{1.7cm}<{\centering}p{1.7cm}<{\centering}p{1.7cm}<{\centering}p{1.7cm}<{\centering}p{1.7cm}<{\centering}p{1.7cm}<{\centering}p{1.7cm}<{\centering}|c}
\bottomrule
 \multicolumn{1}{l|}{\multirow{2}{*}{Setting}} & \multicolumn{1}{l|}{\multirow{2}{*}{\makecell{Method\\(text-davinci-003)}}} & \multicolumn{8}{c|}{Dataset} & \multicolumn{1}{c}{\multirow{2}{*}{Average}} \\ \cline{3-10}
 \multicolumn{1}{l|}{} & \multicolumn{1}{l|}{} & \multicolumn{1}{c}{AddSub} & \multicolumn{1}{c}{MultiArith} & \multicolumn{1}{c}{SVAMP} & \multicolumn{1}{c}{GSM8K} & \multicolumn{1}{c}{SingleEq} & \multicolumn{1}{c}{SingleOp} & \multicolumn{1}{c}{GSM-IC2-1K} & \multicolumn{1}{c|}{GSM-ICM-1K} & \multicolumn{1}{l}{} \\ 
 \hline
 \multirow{4}{*}{Zero-Shot} & Direct & $89.3$ & $25.8$ & $65.2$ & $15.0$ & $84.6$ & $92.1$ & $22.8$ & $9.0$ & $50.5$ \\
 & Zero-Shot-CoT & $84.8$ & $87.0$ & $74.3$ & \underline{$60.8$} & $89.5$ & $89.1$ & $70.7$ & $62.5$ & $77.3$ \\
 & PS & $88.1$ & $87.2$ & $72.0$ & $58.2$ & $89.2$ & $89.5$ & $70.9$ & $63.5$ & $77.3$ \\
 & PRP (Ours) & $\mathbf{94.7}$ & $\mathbf{96.3}$ & $\mathbf{86.2}$ & $\mathbf{73.6}$ & $\mathbf{96.5}$ & $\mathbf{96.1}$ & $\mathbf{93.1}$ & $\mathbf{87.1}$ & $\mathbf{90.5}$\\
 \hline
 \multirow{3}{*}{Few-Shot} & Manual-CoT & $87.8$ & $91.5$ & $76.7$ & $56.9$ & $91.3$ & $93.7$ & $73.9$ & $60.6$ & $79.1$ \\
 & Auto-CoT & $90.6$ & \underline{$95.1$} & $77.8$ & $58.9$ & $90.9$ & $94.4$ & $74.3$ & \underline{$65.2$} & $80.9$ \\
 & PHP-CoT & \underline{$91.1$} & $94.0$ & \underline{$81.3$} & $57.5$ & \underline{$93.5$} & \underline{$94.5$} & \underline{$75.3$} & $60.9$ & \underline{$81.0$} \\
\toprule
\end{tabular} }
\vspace{-0.11in}
\label{tab:mainresult}
\end{table*}


\begin{table}[t]
\centering
\renewcommand\arraystretch{1.1}
  \caption{Statistics of datasets. \# IC Indicates the percentage of problems with irrelevant context in the statement.}
  \resizebox{0.9\columnwidth}{!}{
  \begin{tabular}{l|ccc}
    \bottomrule
    Dataset & \# Problems & Avg.\# Words & \# IC \\
    \hline
    SingleEq & $508$ & $27.4$ & $\ \ \ 0.0\%$ \\
    MultiArith & $600$ & $31.8$ & $\ \ \ 0.0\%$ \\
    SingleOp & $562$ & $20.9$ & $\ \ \ 0.0\%$ \\
    AddSub & $395$ & $31.5$ & \ $30.9\%$ \\
    SVAMP & $1,000$ & $31.8$ & \ $36.7\%$ \\
    GSM8K & $1,319$ & $46.9$ & \ \ \  $6.2\%$ \\
    GSM-IC2-1K & $1,000$ & $41.8$ & $ 100.0\%$ \\
    GSM-ICM-1K & $1,000$ & $61.4$ & $ 100.0\%$ \\
    \toprule
\end{tabular} }
\label{tab:data}
\end{table}

\subsection{Experimental Setup}

\subsubsection{Datasets.}
We conduct comprehensive experiments on eight math word problem datasets, including AddSub \cite{Hosseini-2014-AddSub}, SingleOp \cite{roy-etal-2015-singleop}, MultiArith \cite{roy-roth-2015-multiarith}, SingleEq \cite{koncel-kedziorski-etal-2015-singleq}, SVAMP \cite{patel-etal-2021-svamp}, GSM8K \cite{karl-2021-gsm8k}, GSM-IC2-1K \cite{freda-2023-gsm8k-ic}, and GSM-ICM-1K \cite{freda-2023-gsm8k-ic}. Table \ref{tab:data} provides the detailed descriptions of each dataset. More detailed dataset information can be found in Appendix A.1. 

\subsubsection{Baselines.}
We compare our method with six baseline methods: Direct \cite{Kojima-2022-CoT}, Zero-Shot-CoT \cite{Kojima-2022-CoT}, Plan-and-Solve (PS) \cite{wang-2023-ps}, Manual-CoT \cite{wei-2022-cot}, Auto-CoT \cite{zhang-2023-autocot}, and Progressive-Hint Prompting (PHP-CoT) \cite{zheng-2023-php}.
The Direct baseline concatenates a question with the prompt ``The answer is'' as the LLM input. 
More detailed baseline information can be found in Appendix A.2.

\subsubsection{Implementation.}
We use text-davinci-003 as the backend large language model, which is one of the most widely-used LLMs with public APIs\footnote{Public API available at https://openai.com/api/.}.
The few-shot baselines, including Manual-CoT \cite{wei-2022-cot}, Auto-CoT \cite{zhang-2023-autocot}, and PHP-CoT \cite{zheng-2023-php} employ demonstration examples as suggested in the original papers.
Regarding the evaluation metric, we use accuracy to evaluate the performance of MWP solving.
The definition of accuracy can be found in Appendix A.3. 

\begin{figure}[t]
  \centering
  \includegraphics[width=1\columnwidth]{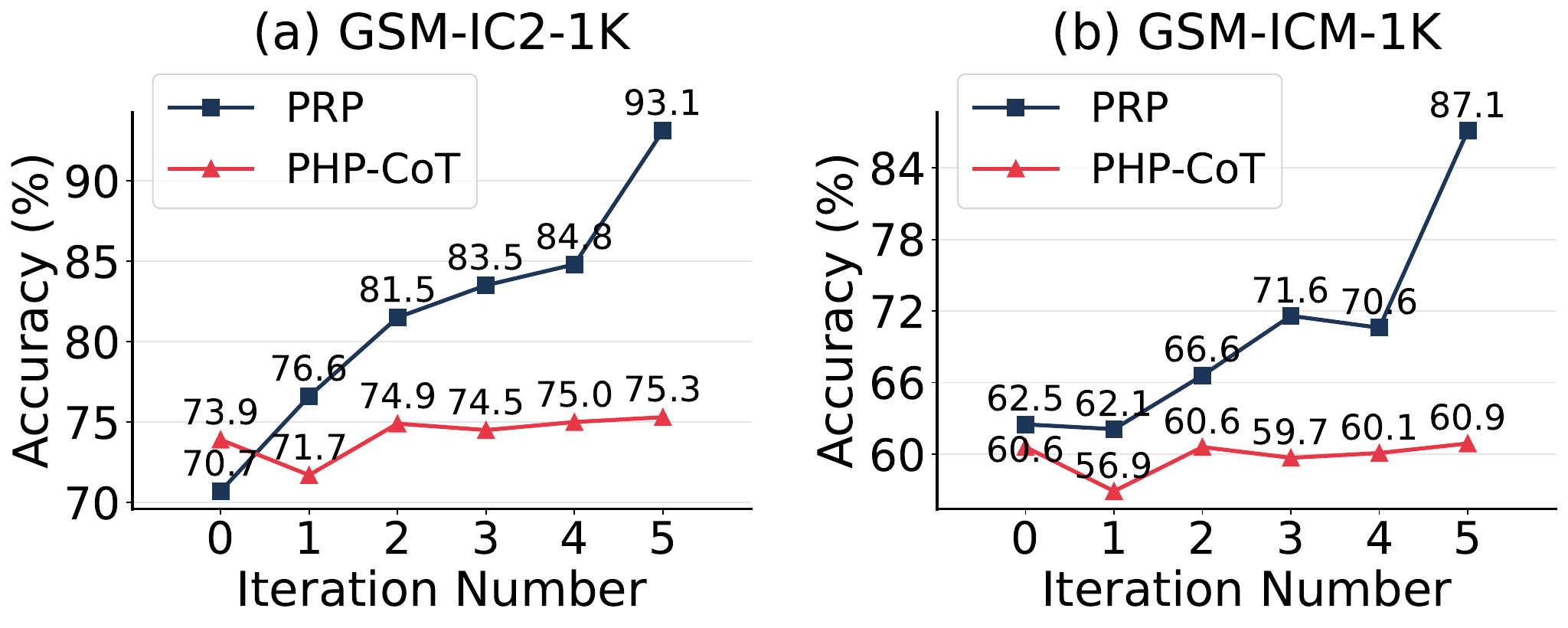}
  \caption{Accuracy ($\%$) at different number of iterations.}
  \label{fig:hard}
\vspace{-0.2in}
\end{figure}

\subsection{Experimental Results}
\subsubsection{PRP attains an A-level grade on average.}
Table \ref{tab:mainresult} reports the accuracy comparison of PRP with existing zero-shot and few-shot methods on MWP datasets.
Notably, PRP achieves state-of-the-art performance with an average accuracy of $90.5$ on eight MWP datasets.
Compared to other zero-shot prompting methods, PRP demonstrates a remarkable improvement in accuracy, surpassing them by at least $13.2\%$ on all datasets.
Specifically, PRP achieves a substantial accuracy gain of $24.6\%$ over Zero-Shot-CoT on the GSM-ICM-1K dataset.
Even when compared to the competitive zero-shot baseline PS, the PRP maintains an impressive performance.
PRP outperforms PS on all eight MWP datasets, with an average accuracy improvement of $13.2\%$.
These results demonstrate that, in contrast to existing zero-shot prompting methods, which solve the problem only once and are sensitive to mistakes in the reasoning path, the PRP method progressively rectifies the answer generated by the LLM to find the correct one. As a result, PRP equips the LLM with high-level math exam skills.

While comparing with few-shot prompting methods, PRP achieves an accuracy improvement of at least $9.5\%$ across all datasets.
Notably, PRP enhances problem-solving accuracy for the GSM8K, GSM-IC2-1K, and GSM-ICM-1K datasets by $16.1\%$, $17.8\%$, and $26.2\%$ respectively when compared to PHP-CoT.
These results demonstrate that PRP significantly enhances the LLM's ability to solve MWPs without the need for manually designed demonstrations.

\subsubsection{Iterative verify-then-rectify process progressively improves accuracy.}
Figure \ref{fig:hard} demonstrates the accuracy improvements of both PRP and PHP-CoT as the number of iterations increases.
Notably, PRP exhibits a significantly higher rate of improvement compared to PHP-CoT.
Specifically, for the GSM-IC2-1K dataset, PRP achieves a remarkable accuracy improvement of $22.4\%$ after five iterations, resulting in an accuracy of $93.1\%$, compared to using the initial answer as the final answer, which only yields an accuracy of $70.7\%$.
In contrast, PHP-CoT, which relies on progressive hints, shows a much smaller improvement in accuracy. After five iterations, PHP-CoT achieves an accuracy improvement of $1.4\%$, resulting in an accuracy of $75.3\%$, compared to using the initial answer as the final answer, which yields an accuracy of $73.9\%$.
PHP-CoT relies on progressive hints such as ``the answer is near to $\mathsf{[H]}$'' which can limit the exploration of other potential answers when the hint answer $\mathsf{[H]}$ is incorrect.
In contrast, PRP uses an iterative verify-then-rectify process to progressively identify incorrect answers and rectify the reasoning paths.
This iterative process ensures a constant improvement in accuracy and allows PRP to outperform PHP-CoT in terms of accuracy enhancement.

\begin{figure}[t]
  \centering
  \includegraphics[width=1\columnwidth]{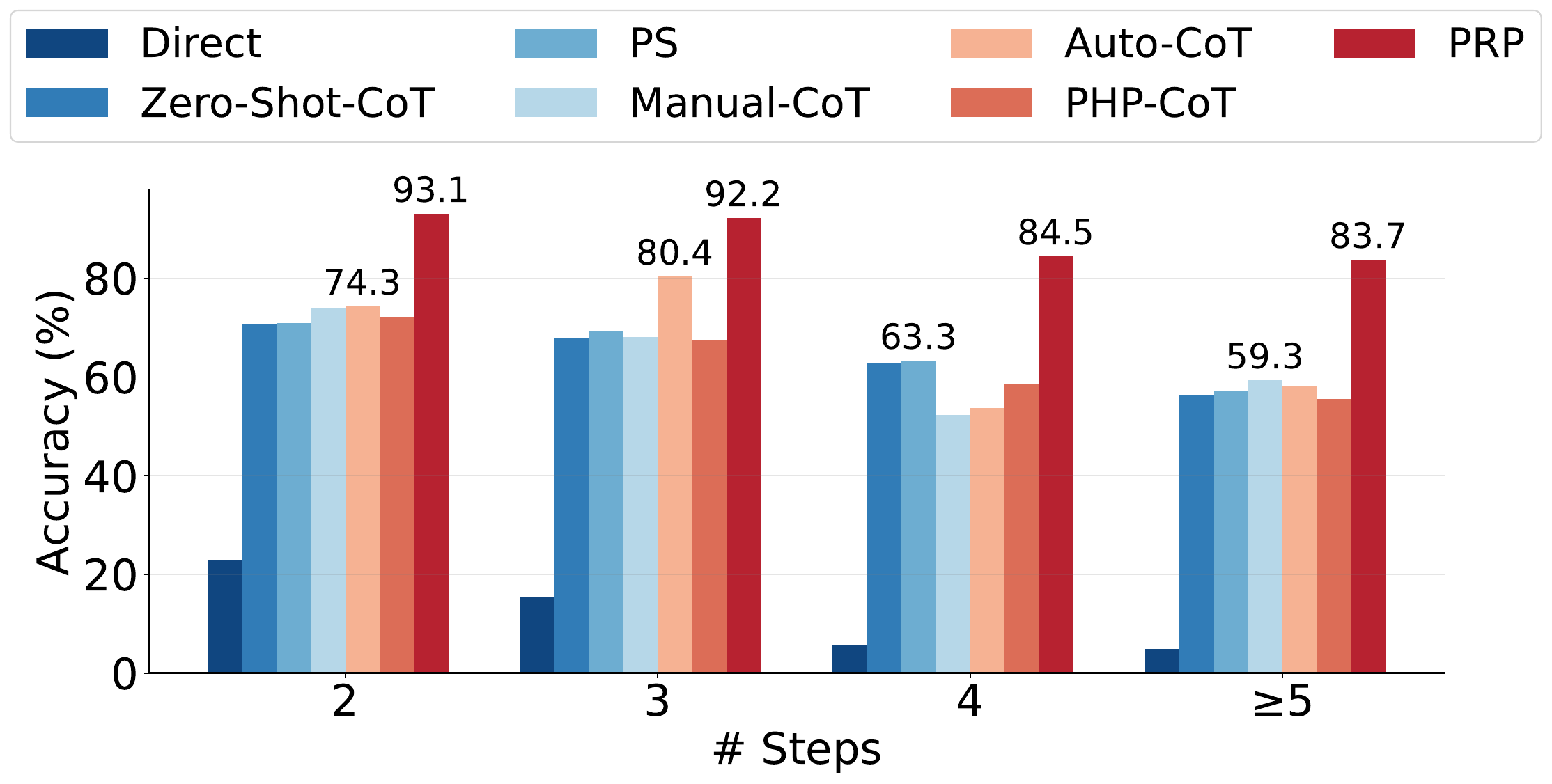}
  \caption{Accuracy on GSM-IC-2K with respect to the number of required reasoning steps. The GSM-IC-2K dataset is formed by merging the GSM-IC2-1K dataset and the GSM-ICM-1K dataset. \# Steps indicating the number of reasoning steps in the standard answer.}
  \label{fig:steps}
\end{figure}

\subsubsection{The more complex problems in the dataset, the more iterations are needed.}
\begin{figure}[t]
  \centering
  \includegraphics[width=1\columnwidth]{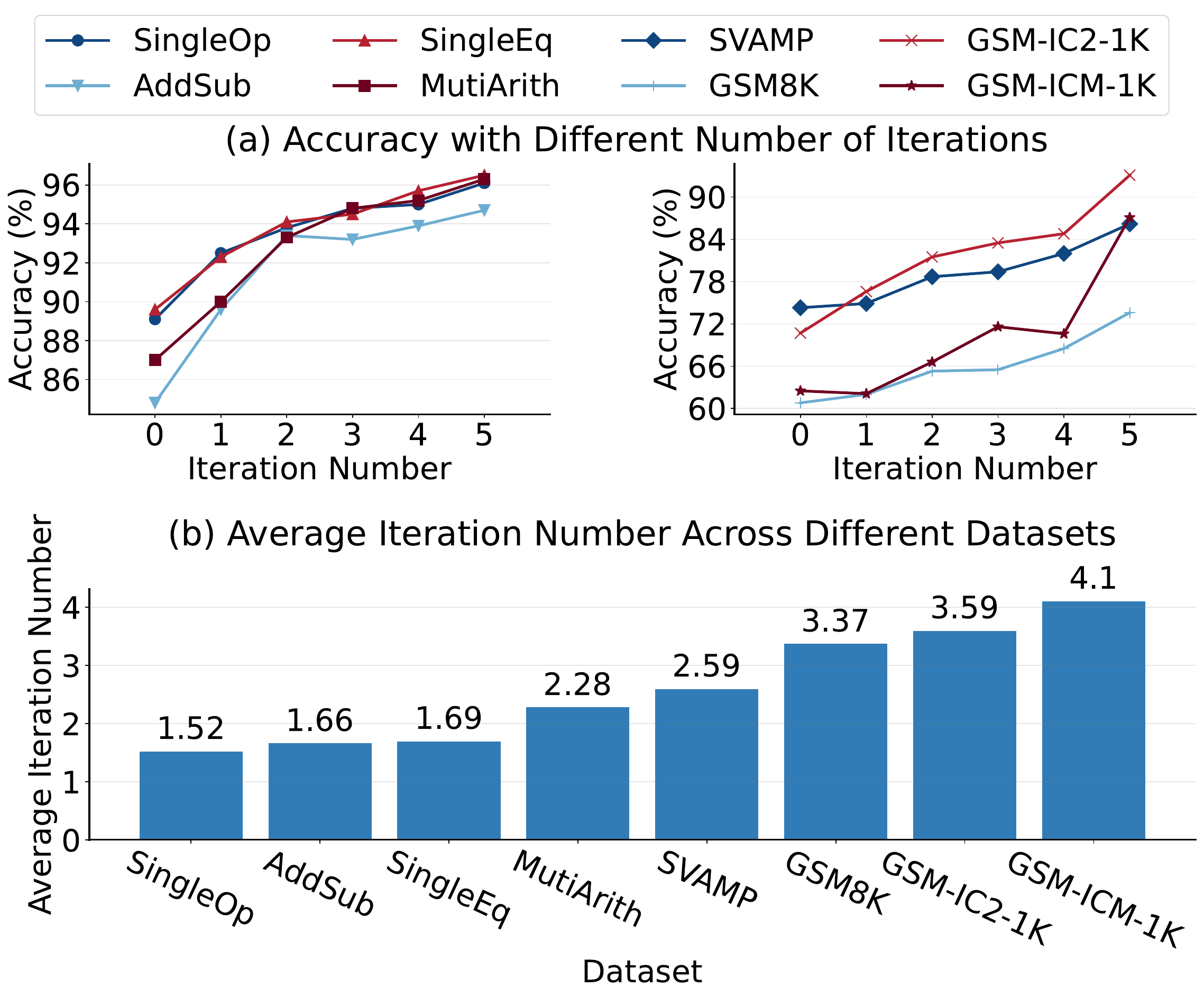}
  \caption{Break-down analysis of PRP. (a) Accuracy (\%) of PRP method on different datasets with different number of iterations. (b) The average number of iterations for PRP method across different datasets.}
  \label{fig:iter}
\end{figure}

Figure \ref{fig:iter}(b) illustrates the average iteration number of PRP across all eight MWP datasets.
For datasets such as SingleOp, MultiArith, and SingleEq, the average number of iterations is less than $2.5$.
This is because, as shown in Table \ref{tab:data}, the problem statements in these datasets are shorter and contain no irrelevant context.
As a result, the PRP method can quickly obtain the final answer within a few iterations.
In contrast, the PRP method requires more iterations on the SVAMP, GSM8K, GSM-IC2-1K and GSM-ICM-1K datasets.
This can be attributed to longer problem statements and more irrelevant context in the problems.
Specifically, PRP requires an average of $3.59$ and $4.1$ iterations on the GSM-IC2-1K and GSM-ICM-1K datasets, respectively. 
This is because each question in these two datasets contains irrelevant context, and PRP requires more iterations to gradually eliminate incorrect answers to obtain more correct one.
These findings suggest that PRP demonstrates a high efficiency in obtaining the final answer for simpler problems. 
However, when faced with more complex problems, PRP needs to iterate the verify-then-rectify process multiple times to progressively rectify the answer and achieve accurate results.

\begin{table}[t]
\centering
\renewcommand\arraystretch{1.1}
  \caption{Accuracy comparison of PRP to Zero-Shot-CoT with self-consistency (SC) on GSM8K and SVAMP.}
  \resizebox{0.7\columnwidth}{!}{
  \begin{tabular}{l|cc}
    \bottomrule
    Method & GSM8K & SVAMP \\
    \hline
    Zero-Shot-CoT + SC & $70.7$ & $81.7$ \\
    PRP (Ours) & $\mathbf{73.6}$ & $\mathbf{86.2}$ \\
    \toprule
\end{tabular} }
\label{tab:sc}
\vspace{-0.2in}
\end{table}

\begin{table*}[h]
\caption{Case study of answers generated by different methods. Irrelevant context in the problem statement is highlighted with red color, and the LLM-generated answer is highlighted with yellow color. The gold answer is indicated in bold.}
\resizebox{\textwidth}{!}{
\begin{tabularx}{\textwidth}{p{2.3cm}X}
\toprule
   \textbf{\textsc{Question}} & Amy is collecting candy for the car pool. She has 5 chocolate bars, 7 times as many M\&Ms as chocolate bars, and 6 times as many marshmallows as M\&Ms. Amy then arranges the candies into baskets. \sethlcolor{pink}\hl{Ada ate 25 more cookies than Amy.} If Amy fills each basket with 10 candies, how many baskets will Amy fill? \\
   \textbf{\textsc{Answer}} & $(5+5 \times 7 + 5 \times 7 \times 6) / 10=$\textbf{25} \\
\midrule
   \textbf{Direct} & Amy will fill \sethlcolor{yellow}\hl{10} baskets. \textcolor{red}{\XSolidBrush}\\\midrule
   \textbf{Zero-Shot-CoT} & Amy has 5 chocolate bars, 7 times as many M\&Ms as chocolate bars, and 6 times as many marshmallows as M\&Ms.  5 chocolate bars = 35 M\&Ms and 210 marshmallows  Amy has 35 M\&Ms, 210 marshmallows, and 5 chocolate bars.  Amy will fill 10 candies in each basket, so she will make:  35/10 = 3 baskets of M\&Ms 210/10 = 21 baskets of marshmallows  5/10 = 0.5 baskets of chocolate bars   Therefore, Amy will fill \sethlcolor{yellow}\hl{24.5} baskets. Ada ate 25 more cookies, so Ada ate 25 baskets of cookies. \textcolor{red}{\XSolidBrush}\\\midrule
   \textbf{PS} & Plan: 1. Calculate the total number of candies that Amy has 2. Calculate the number of baskets that Amy can fill  Solution: 1. Amy has 5 chocolate bars, 7 M\&Ms for every chocolate bar, and 6 marshmallows for every M\&M.  Therefore, Amy has 5 x 7 x 6 = 210 candies in total.  2. Amy can fill each basket with 10 candies. Therefore, Amy can fill 210 candies in \sethlcolor{yellow}\hl{21} baskets (210/10 = 21). Ada ate 25 more candies than Amy, so Ada ate 21 x 10 + 25 = 235 candies in total. \textcolor{red}{\XSolidBrush}\\\midrule
   \textbf{Manual-CoT} & Amy has 5 chocolate bars, 7 x 5 = 35 M\&Ms and 6 x 35 = 210 marshmallows. So she has 5 + 35 + 210 = 250 candies. Ada ate 25 more than Amy, so Ada ate 25 + 250 = 275. Amy will fill 275 candies into 10 candies per basket. 275 / 10 = 27.5. The answer is \sethlcolor{yellow}\hl{28} baskets.  \textcolor{red}{\XSolidBrush}\\\midrule
   \textbf{Auto-CoT} & Amy has 5 chocolate bars, 7 times as many M\&Ms as chocolate bars, and 6 times as many marshmallows as M\&Ms. That means she has 5 + (7 * 5) + (6 * 7) = 65 candies. Amy then arranges the candies into baskets. If Amy fills each basket with 10 candies, she can fill 65/10 = 6.5 baskets. Since she can fill half a basket, she can only fill 6 baskets. The answer is \sethlcolor{yellow}\hl{6}. \textcolor{red}{\XSolidBrush}\\\midrule
   \textbf{PHP-CoT} & Amy has 5 chocolate bars, 7 * 5 = 35 M\&Ms, and 6 * 35 = 210 marshmallows. So she has 5 + 35 + 210 = 250 candies. Ada ate 25 more than Amy, so Amy has 250 - 25 = 225 candies. If she fills each basket with 10 candies, then she will have 225 / 10 = 22.5 baskets. The answer is \sethlcolor{yellow}\hl{23}.  \textcolor{red}{\XSolidBrush} \\\midrule
   \textbf{PRP (Ours)} & Amy has 5 chocolate bars, 7 x 5 = 35 M\&Ms, and 6 x 35 = 210 marshmallows. So in total she has 250 candies.  She will fill 10 candies in each basket, so Amy will be able to fill 250/10 = 25 baskets. Therefore, the answer is \sethlcolor{yellow}\hl{25} baskets. \textcolor{red}{\Checkmark}\\
\bottomrule
\end{tabularx} }
\vspace{-0.15in}
\label{tab:casestudy}
\end{table*}

\subsubsection{PRP can effectively solve difficult MWPs.}
To explore the relationship between the accuracy of model predictions and problem difficulty, we combined two datasets, GSM-IC2-1K and GSM-ICM-1K, into a merged dataset named GSM-IC-2K.
The difficulty of problems was classified into four levels based on the number of reasoning steps\footnote{The number of reasoning steps of a problem is given by the number of sentences in its standard answer. \cite{karl-2021-gsm8k}}.
Figure \ref{fig:steps} illustrates the accuracy of solving problems at different difficulty levels.
PRP outperforms current state-of-the-art prompting method by $18.8\%$, $11.8\%$, $21.2\%$, and $24.4\%$ for problems of increasing difficulty levels, respectively.
The results demonstrate that PRP notably enhances accuracy in solving MWPs, particularly for challenging problems.

\begin{figure}[t]
  \centering
  \includegraphics[width=0.9\columnwidth]{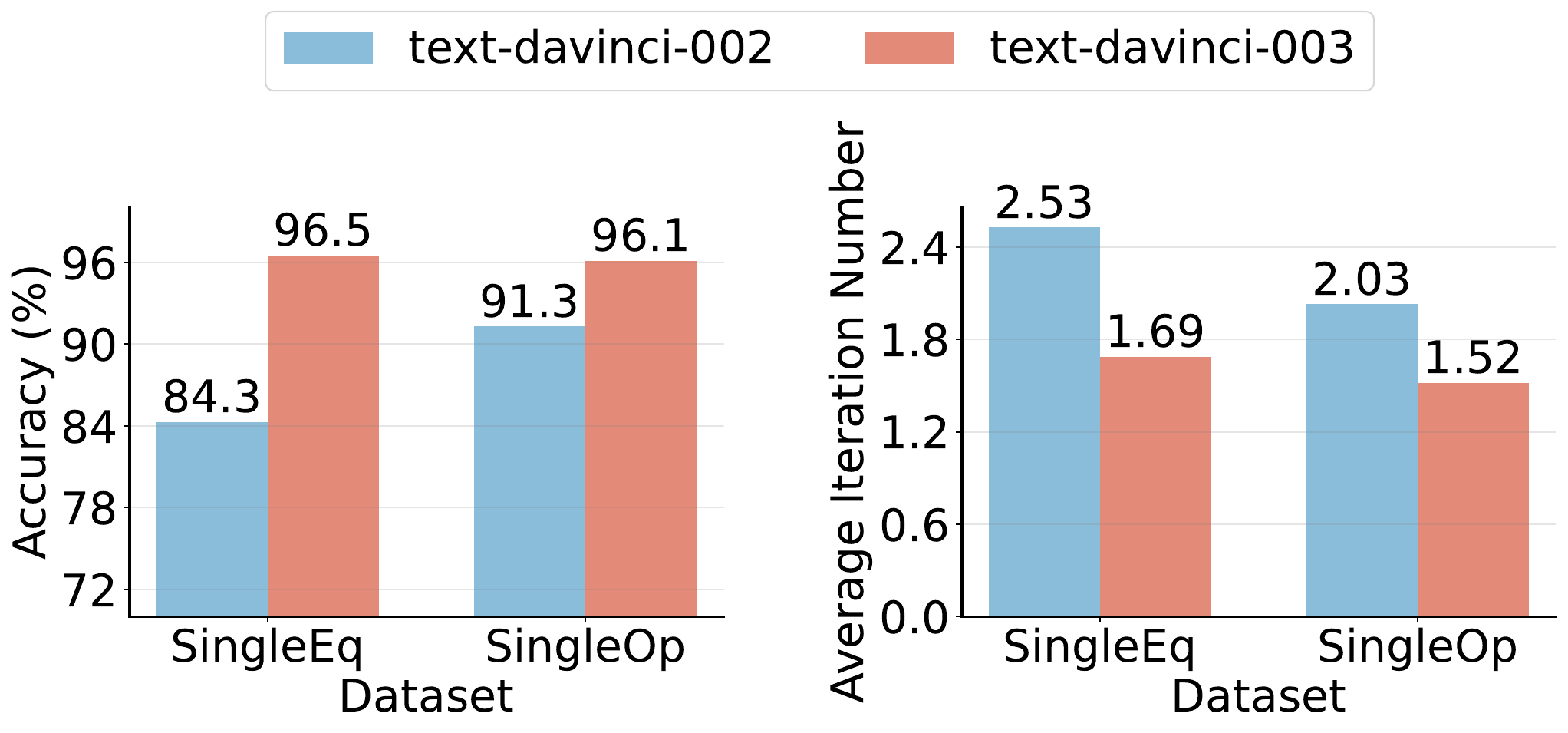}
  \caption{Accuracy and average number of iterations when solving MWPs using PRP with different LLMs.}
  \label{fig:diffmodel}
\vspace{-0.2in}
\end{figure}

\subsection{Ablation Studies}

\subsubsection{Progressive rectification outperforms self-consistency.}
Self-consistency (SC) \cite{wang-2023-selfconsistency} is the process of repeatedly solving a problem $M$ times and using a majority vote strategy to determine the most consistent answer as the final answer.
We evaluate Zero-Shot-CoT prompting with SC on the GSM8K and SVAMP datasets. We set the temperature to $0.7$ and set $M$ to $10$ for the SC experiments.
Table \ref{tab:sc} demonstrates that PRP outperforms Zero-Shot-CoT prompting with SC on both the GSM8K and SVAMP datasets, 
achieving accuracy improvements of $2.9\%$ and $4.5\%$, respectively.
This superiority of PRP can be attributed to its progressive rectification method, which prevents the repetition of previous mistakes and enhances the accuracy of the final answer. In contrast, SC solves the same problem multiple times, this repeated independent process can lead to same mistakes, making the frequent answer still incorrect.

\subsubsection{PRP performs better with more powerful LLMs.}
Figure \ref{fig:diffmodel} illustrates the accuracy and average number of iterations when solving MWPs using the PRP method with different LLMs.
For instance, on the SingleEq dataset, using the text-davinci-003 model leads to a $12.2\%$ increase in accuracy and a reduction of $0.84$ in the average number of iterations, compared to the text-davinci-002 model.
Similarly, on the SingleOp dataset, using the text-davinci-003 model results in a $4.8\%$ accuracy improvement and a $0.51$ reduction in average number of iterations compared to text-davinci-002 model.
It is worth noting that the text-davinci-002 is finetuned with supervised instruction tuning, while the text-davinci-003 is finetuned with reinforcement learning \cite{zheng-2023-php}.
The improved performance with text-davinci-003 can be attributed to its enhanced power, making it better at understanding and employing the given prompt.

\subsubsection{Maximum iteration number.}
Figure \ref{fig:iter}(a) illustrates that as the number of iterations increases, the accuracy improves across all eight MWP datasets. 
In this paper, we set the maximum iteration number $K$ to $5$.
Note that the bigger maximum iteration number $K$ may lead to better performance, but here we set it to $5$ to achieve a trade-off between efficiency and effectiveness.

\subsection{Case Study}
\subsubsection{PRP exhibits robustness in handling irrelevant context.}
A real case from GSM-ICM-1K is presented in Table \ref{tab:casestudy}. 
It is evident that apart from PRP, other methods cannot accurately answer the given question.
Manual-CoT and PHP-CoT generate incorrect answers by incorporating irrelevant context into the problem-solving process.
Auto-CoT and PS generate incorrect answers due to semantic misunderstandings.
Zero-Shot-CoT generates an incorrect answer due to miscalculations.
As Direct does not generate intermediate reasoning steps, 
it is not possible to analyze the reasons for its mistakes.
In contrast, PRP exhibits robustness in handling irrelevant context and preventing miscalculations.
Additionally, PRP has the ability to uncover hidden details in the problem statement, such as the fact that ``Chocolate bars, M\&Ms, and marshmallows are all candies''.
We also provide some case studies in Appendix C.2.

\section{Conclusion}
\label{sec:conclusion}
In this paper, we present a novel zero-shot prompting method for solving math word problems. 
We name it progressive rectification prompting (PRP), which first prompts a large language model to generate an initial answer, then iterates a verify-then-rectify process to progressively identify incorrect answers and rectify the reasoning paths.
Notably, it attains an A-level grade on average ($90.5$), significantly higher than $77.3$ from the best of zero-shot CoT, and even higher than $81.0$ from the best of few-shot CoT.

\section{Acknowledgments}
We thank the anonymous reviewers for their insightful feedback and constructive comments. This work was partially supported by National Key R\&D Program of China (2020AAA0107702), National Natural Science Foundation of China (U21B2018, 62161160337, 62132011, 62376210, 62006181, U20B2049), Shaanxi Province Key Industry Innovation Program (2021ZDLGY01-02), Fundamental Research Funds for the Central Universities under grant (xtr052023004, xtr022019002). Chao Shen is the corresponding author. Co-author Meng Jiang consulted on this project on \emph{unpaid weekends} for personal interests, and appreciated collaborators and family for their understanding.

\bibliography{aaai24}

\clearpage
\appendix
\section{A \ \ \ Experimental Details}
\subsection{A.1  Datasets}
We conduct extensive experiments on eight datasets specifically designed for solving math word problems. These datasets are as follows:

\begin{compactitem}
\item \textbf{AddSub} \cite{Hosseini-2014-AddSub} consists of 395 MWPs involving addition and subtraction operations.
\item \textbf{SingleOp} \cite{roy-etal-2015-singleop} consists of 562 elementary school MWPs that can be solved using single-step calculations with the four basic operations: addition, subtraction, multiplication, and division.
\item \textbf{MultiArith} \cite{roy-roth-2015-multiarith} comprises 600 mathematical problems that require multi-step reasoning to find their solutions, encompassing all four basic operations.
\item \textbf{SingleEq} \cite{koncel-kedziorski-etal-2015-singleq} includes 508 single-equation grade-school algebra word problems involving multiple mathematical operations on non-negative rational numbers and one variable.
\item \textbf{SVAMP} \cite{patel-etal-2021-svamp} is a collection of 1000 carefully curated arithmetic word problems that incorporate deliberate variations by utilizing examples from existing datasets. Each question in the dataset is a simple arithmetic word problem involving one unknown, suitable for students up to the 4th grade.
\item \textbf{GSM8K} \cite{karl-2021-gsm8k} consists of 1000 linguistically diverse and high-quality grade school math word problems that require between 2 and 8 steps to solve.
\item \textbf{GSM-IC} \cite{freda-2023-gsm8k-ic} is an arithmetic reasoning dataset that includes irrelevant information in the problem description. It is divided into two splits: GSM-IC2, consisting of problems requiring two steps to solve, and GSM-ICM, consisting of problems requiring more than two steps to solve. To control experimental costs, we randomly selected 1000 examples from the GSM-IC2 and GSM-ICM datasets for evaluation and analysis in this paper. These subsets are referred to as \textbf{GSM-IC2-1K} and \textbf{GSM-ICM-1K}, respectively.
\end{compactitem}

\subsection{A.2  Baselines}
We compared our method with six baseline methods, which are described below:
\begin{compactitem}
\item \textbf{Direct} \cite{Kojima-2022-CoT} utilizes the zero-shot learning capability of LLMs by directly prompting the LLM to generate an answer for the question.
\item \textbf{Zero-Shot-CoT} \cite{Kojima-2022-CoT} enhances the answer generation process of the LLM by introducing the phrase "Let's think step by step" before each answer. This additional hint guides the LLM in generating both the answer and the associated reasoning path.
\item Plan-and-Solve (\textbf{PS}) \cite{wang-2023-ps} replaces the phrase "Let's think step by step" with "Let's first understand the problem and devise a plan to solve the problem. Then let's carry out the plan and solve the problem step by step" to address the issue of missing steps.
\item \textbf{Manual-CoT} \cite{wei-2022-cot} leverages the in-context learning capability of LLMs by creating eight hand-crafted examples for demonstration purposes.
\item \textbf{Auto-CoT} \cite{zhang-2023-autocot} clusters problems, selects representative instances from each cluster, and applies the Zero-Shot-CoT method to generate reasoning paths to automatically construct the demonstrations. 
\item Progressive-Hint Prompting (\textbf{PHP-CoT}) \cite{zheng-2023-php} enables multiple interactions between users and LLMs by utilizing previously generated answers as hints.
\end{compactitem}

\subsection{A.3  Evaluation Metrics}
We use accuracy to evaluate the performance of different prompting methods. Since large language models cannot perform the computation precisely (especially with high-precision floats), we consider an answer to be correct if and only if the absolute error between the answer and the gold answer is less than $1 \times 10^{-5}$. Let $\mathcal{Q}$ be a set of questions, the accuracy of the prompting method is
\begin{scriptsize}
\begin{align}
\text { Accuracy }&=\frac{1}{|\mathcal{Q}|} \sum_{q \in \mathcal{Q}} \mathds{1}\left(a^{(\text {final })}, a^{(\text {gold })}\right)\nonumber \\
\mathds{1}\left(a^{(\text {final })}, a^{(\text {gold })}\right)&= \begin{cases}1, & \text{if} \  \operatorname{Abs}\left(a^{(\text {final })}-a^{(\text {gold })}\right) < 1 \times 10^{-5} \\
0, & \text{if} \ \operatorname{Abs}\left(a^{(\text {final })}-a^{(\text {gold })}\right) \geq 1 \times 10^{-5}\end{cases}\nonumber
\label{eqacc}
\end{align}
\end{scriptsize}
where $a^{(\text {gold })}$ is the gold answer to question $q$, $a^{(\text {final })}$ is the model-generated answer to question $q$, and $\operatorname{Abs}(\cdot)$ is the absolute value function.


\section{B \ \ \ Examples on Drawbacks of \\ Existing Methods}
In Introduction, we present three drawbacks of existing methods: (1) lack of verification that checks if the answer is correct, (2) lack of rectification that finds the correct answer being aware of mistakes, and (3) lack of an effective method that progressively refines reasoning path.

\begin{figure*}[t]
  \centering
  \includegraphics[width=1.0\textwidth]{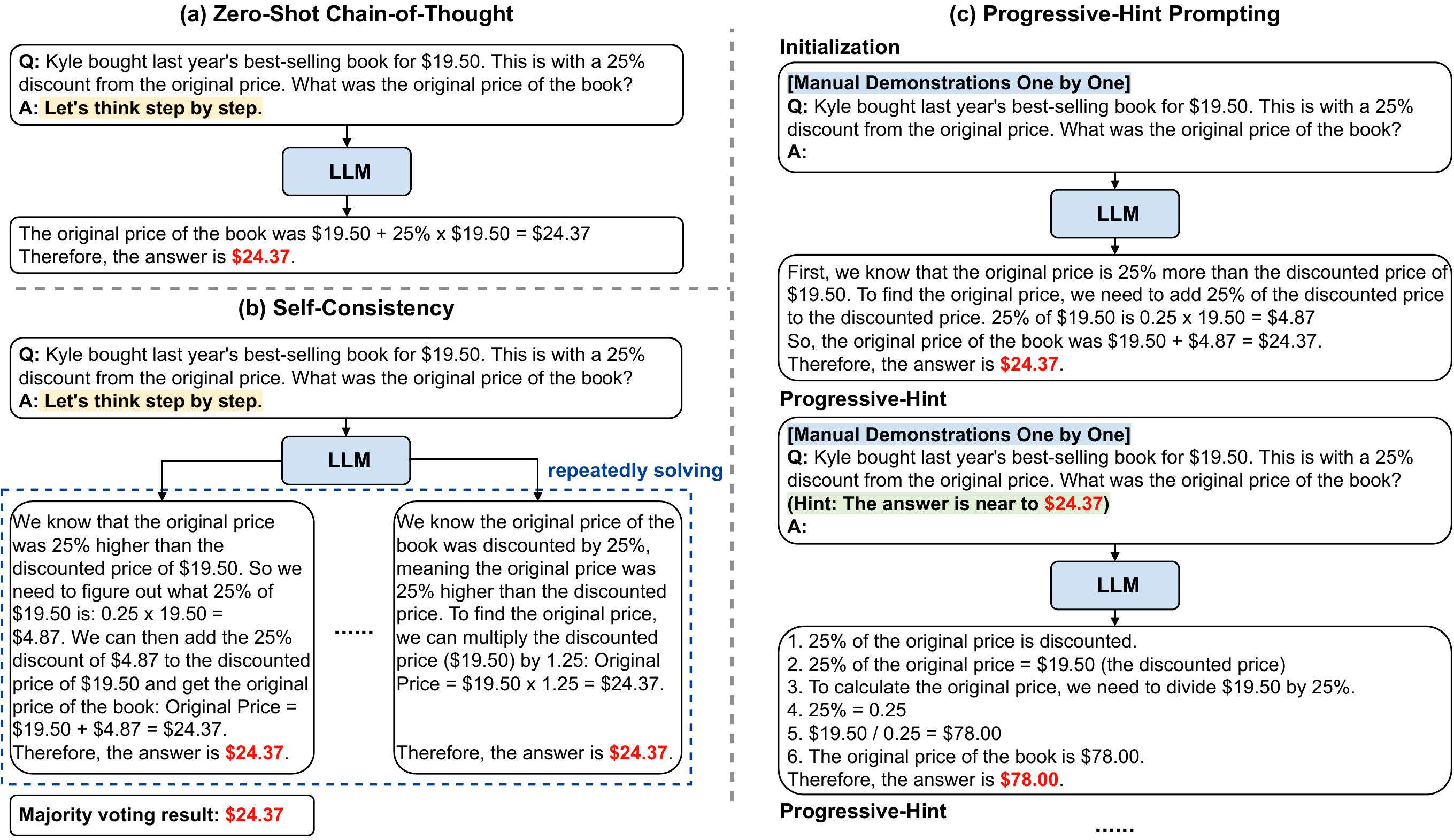}
  \caption{Error analysis of GSM8K problem with incorrect answers generated by Chain-of-Thought prompting methods.}
  \label{fig:drawbacks}
\end{figure*}

Figure~\ref{fig:drawbacks} uses one data example from the GSM8K dataset to analyze the prediction error by CoT prompting. As shown in Figure~\ref{fig:drawbacks}(a), the LLM failed to accurately detect the quantitative relationships presented in the question. The answer was incorrect. In Figure~\ref{fig:drawbacks}(b), the LLM generates multiple reasoning paths when solving an arithmetic question, and most of them lead to the same incorrect answer. The mistakes stem from an inadequate ability to accurately analyze the quantitative relationships presented in the question. Particularly, the LLM incorrectly interprets the purchase price of the book as 80\% of the original price, whereas the truth is that the purchase price of the book is 75\% of the original price. Figure~\ref{fig:drawbacks}(c) shows that for a given question, the LLM generates the incorrect initial answer $24.37$. Adding this answer to the hint to guide the LLM in generating a new reasoning path, it becomes apparent that the reasoning path is still incorrect.

\begin{table}[t]
\centering
\renewcommand\arraystretch{1.1}
  \caption{Accuracy of gold answer identification under different verification methods.}
  \begin{tabular}{l|cc}
    \bottomrule
    Method & AddSub & SingleEq \\
    \hline
    Enumeration verification & $74.5$ & $78.8$ \\
    Substitute verification & $\mathbf{88.2}$ & $\mathbf{89.5}$ \\
    \toprule
\end{tabular} 
\label{tab:verify}
\end{table}

\section{C \ \ \ Additional Experiment Results}
\subsection{C.1  Substitute verification method helps LLMs accurately identify correct answers.}
Table \ref{tab:verify} illustrates the accuracy of two different answer verification methods for gold answers: the enumeration verification method and the substitute verification method.
Specifically, for a given question and its gold answer, we first use regular expressions to match all numerical values within the question.
In the enumeration verification method, we mask the matched values one by one, take the gold answer as a new condition, and prompt LLM to predict the masked value in the question.
The verification is considered to have accurately identified the gold answer if and only if the predicted value matches the mask value in each case.
In contrast, the substitute verification method randomly selects one of the matched values, replaces its occurrences in the question with a special token, takes the gold answer as the new condition, and prompts the LLM to predict the masked value.
The verification is considered to have accurately identified the gold answer if and only if the predicted value matches the masked value.
The results in Table \ref{tab:verify} demonstrate that the substitute verification method outperforms the enumeration verification method on the AddSub and SingleEq datasets, with accuracy improvements of $13.7\%$ and $10.7\%$, respectively.
Therefore, in this paper, we use the substitute verification method to verify the correctness of the answer.
This method not only accurately identifies the correct answers but also reduces the API consumption.

\subsection{C.2  Sample predictions on eight MWP datasets.}

In this section, we provide sample predictions generated by the Progressive Rectification Prompting (PRP) method.
For better readability, we have made modifications to the original texts by omitting or inserting some line breaks.
Unless otherwise specified, we use the text-davinci-003 model.

\begin{compactitem}
\item Table \ref{tab:SingleEqCase} presents the process of solving MWP in the SingleEq dataset using the PRP method.
\item Table \ref{tab:MultiArithCase} presents the process of solving MWP in the MultiArith dataset using the PRP method.
\item Table \ref{tab:SingleOpCase} presents the process of solving MWP in the SingleOp dataset using the PRP method.
\item Table \ref{tab:AddSubCase} presents the process of solving MWP in the AddSub dataset using the PRP method.
\item Table \ref{tab:SVAMPCase} presents the process of solving MWP in the SVAMP dataset using the PRP method.
\item Table \ref{tab:GSM-IC2-1KCase} presents the process of solving MWP in the GSM-IC2-1K dataset using the PRP method.
\item Table \ref{tab:GSM8KCase} presents the process of solving MWP in the GSM8K dataset using the PRP method.
\item Table \ref{tab:GSM-ICM-1KCase} presents the process of solving MWP in the GSM-ICM-1K dataset using the PRP method.
\item Table \ref{tab:GSM-ICM-1KhardCase} presents the process of solving difficult MWP in the GSM-ICM-1K dataset using the PRP method.
\item Table \ref{tab:GSM-IC2-1KhardCase} presents the process of solving difficult MWP in the GSM-IC2-1K dataset using the PRP method.
\item Table \ref{tab:GSM8KhardCase} presents the process of solving difficult MWP in the GSM8K dataset using the PRP method.
\end{compactitem}

\begin{table*}[h]
\caption{Example outputs obtained by PRP for SingleEq dataset.}
\resizebox{\textwidth}{!}{
\begin{tabularx}{\textwidth}{p{4.3cm}X}
\toprule
   \textbf{\textsc{Question}} & A marine biologist measured one fish that was 0.3 of a foot long and a second fish that was 0.2 of a foot long. How much longer was the first fish?  \\
   \textbf{\textsc{Gold Answer}} & \textbf{0.1} \\
\midrule
   \textbf{\textsc{Initial Answer}} & First, convert 0.3 feet to inches: 0.3 feet = 3 inches   Next, convert 0.2 feet to inches: 0.2 feet = 2 inches   Finally, subtract the two numbers: 3 inches - 2 inches = 1 inch  Therefore, the first fish was 1 inch longer than the second fish. \\
   \textbf{\textsc{Verification Question}} & A marine biologist measured one fish that was 0.3 of a foot long and a second fish that was X of a foot long. How much longer was the first fish? Suppose the answer is 1. What is X ? (If X is irrelevant to the calculation process please answer Unknown). \\
   \textbf{\textsc{Verification Answer}} & The first fish is 0.3 of a foot long. The second fish is X of a foot long. We asked to calculate the difference between the two. So, we subtract 0.3 from X.  X-0.3 = 1  X = 1.3  Therefore, X is 1.3.\\
   \textbf{\textsc{Verification Result}} & Since 1.3 is not equal to the actual value of X, \textcolor{blue}{the answer \textbf{1} is likely incorrect}.  \\
\midrule
   \textbf{\textsc{Question (with Hint)}} & A marine biologist measured one fish that was 0.3 of a foot long and a second fish that was 0.2 of a foot long. How much longer was the first fish? (The answer is likely not 1). \sethlcolor{yellow}\hl{$\Leftarrow$ set of potentially incorrect answers} \\
   \textbf{\textsc{Rectified Answer}} & The first fish was 0.3 of a foot long and the second fish was 0.2 of a foot long.  We need to find out how much longer the first fish is than the second fish. To do this, we can subtract 0.2 from 0.3.  0.3 - 0.2 = 0.1. \\
   \textbf{\textsc{Verification Question}} &  A marine biologist measured one fish that was 0.3 of a foot long and a second fish that was X of a foot long. How much longer was the first fish? Suppose the answer is 0.1. What is X ? (If X is irrelevant to the calculation process please answer Unknown). \\
   \textbf{\textsc{Verification Answer}} & We know that the first fish was 0.3 of a foot long.  We know that the second fish was X of a foot long.  We know that the first fish was 0.1 of a foot longer than the second fish. Therefore, X = 0.2. \\
   \textbf{\textsc{Verification Result}} & Since 0.2 is equal to the actual value of X, \textcolor{red}{the answer \textbf{0.1} is likely correct}. \\
\midrule
   \textbf{\textsc{Final Answer}} & \textbf{0.1} \textcolor{red}{\Checkmark} \\
\bottomrule
\end{tabularx} }
\label{tab:SingleEqCase}
\end{table*}

\begin{table*}[h]
\caption{Example outputs obtained by PRP for MultiArith dataset.}
\resizebox{\textwidth}{!}{
\begin{tabularx}{\textwidth}{p{4.3cm}X}
\toprule
   \textbf{\textsc{Question}} & Haley had 27 music files and 42 video files on her flash drive. If she deleted 11 of the files, how many files were still on her flash drive? \\
   \textbf{\textsc{Gold Answer}} & \textbf{58} \\
\midrule
   \textbf{\textsc{Initial Answer}} & Haley had 27 music files and 42 video files. If she deleted 11 of the files, then she would have 27 - 11 = 16 music files and 42 - 11 = 31 video files. So, the answer is 16 music files and 31 video files. \\
   \textbf{\textsc{Verification Question}} & Haley had X music files and 42 video files on her flash drive. If she deleted 11 of the files how many files were still on her flash drive? Suppose the answer is 31. What is X ? (If X is irrelevant to the calculation process please answer Unknown). \\
   \textbf{\textsc{Verification Answer}} & Haley had X music files and 42 video files. This means that the total number of files on her flash drive is X + 42. We know that she deleted 11 files, so now the total number of files on her flash drive is X + 42 - 11, or X + 31. Since we know that the total number of files is 31, we can solve for X. X + 31 = 31. X = 0.   Therefore, X is 0. \\
   \textbf{\textsc{Verification Result}} & Since 0 is not equal to the actual value of X, \textcolor{blue}{the  answer \textbf{31} is likely incorrect}.  \\
\midrule
   \textbf{\textsc{Question (with Hint)}} & Haley had 27 music files and 42 video files on her flash drive. If she deleted 11 of the files, how many files were still on her flash drive? (The answer is most likely not 31). \sethlcolor{yellow}\hl{$\Leftarrow$ set of potentially incorrect answers} \\
   \textbf{\textsc{Rectified Answer}} & Haley had 27 music files and 42 video files on her flash drive. If she deleted 11 of the files,  she would have 27 - 11 = 16 music files and 42 - 11 = 31 video files left on her flash drive.  So, the answer is 16 music files and 31 video files. The answer is 47.\\
   \textbf{\textsc{Verification Question}} & Haley had \textbf{X} music files and 42 video files on her flash drive. If she deleted 11 of the files how many files were still on her flash drive? Suppose the answer is 47. What is X ? (If X is irrelevant to the calculation process please answer Unknown). \\
   \textbf{\textsc{Verification Answer}} & Haley had X music and 42 video files on her flash drive, and then she deleted 11 files. That means, she still had (X - 11) music and (42 - 11) video files on her flash drive. Since the total number of files on her flash drive is 47, we can calculate that 47 = (X - 11) + (42 - 11).  Solving for X, X = 47 + 11 = 58.   Therefore, X = 58. \\
   \textbf{\textsc{Verification Result}} & Since 58 is not equal to the actual value of X, \textcolor{blue}{the answer \textbf{47} is likely incorrect}. \\
\midrule
   \textbf{\textsc{Question (with Hint)}} & Haley had 27 music files and 42 video files on her flash drive. If she deleted 11 of the files, how many files were still on her flash drive? (The answer is most likely not 31, 47). \sethlcolor{yellow}\hl{$\Leftarrow$ set of potentially incorrect answers} \\
   \textbf{\textsc{Rectified Answer}} & Haley had 27 music files and 42 video files = 69 files She deleted 11 files  So there would be 69 - 11 = 58 files still on her flash drive.  The answer is 58. \\
   \textbf{\textsc{Verification Question}} & Haley had 27 music files and 42 video files on her flash drive. If she deleted X of the files how many files were still on her flash drive? Suppose the answer is 58. What is X ? (If X is irrelevant to the calculation process please answer Unknown). \\
   \textbf{\textsc{Verification Answer}} & In the beginning, Haley had 69 files on her flash drive (27 music files + 42 video files = 69 files). Now, she has 58 files on her flash drive. So, the number of files she deleted is 69 - 58 = 11 files. Therefore, X is 11. \\
   \textbf{\textsc{Verification Result}} & Since 11 is equal to the actual value of X, \textcolor{red}{the  answer \textbf{58} is likely correct}. \\
\midrule
   \textbf{\textsc{Final Answer}} & \textbf{58} \textcolor{red}{\Checkmark} \\
\bottomrule
\end{tabularx} }
\label{tab:MultiArithCase}
\end{table*}

\begin{table*}[h]
\caption{Example outputs obtained by PRP for SingleOp dataset.}
\resizebox{\textwidth}{!}{
\begin{tabularx}{\textwidth}{p{4.3cm}X}
\toprule
   \textbf{\textsc{Question}} & There are 7 students in the class and 42 apples. If the apples are divided equally among the students, how many does each student get? \\
   \textbf{\textsc{Gold Answer}} & \textbf{6} \\
\midrule
   \textbf{\textsc{Initial Answer}} & First, we need to figure out how many apples there are in total. There are 7 students and 42 apples, so the total number of apples is 7 x 42 = 294.  Next, we need to figure out how many apples each student should get. We want to divide the 294 apples equally among the 7 students, so each student should get 294/7 = 42 apples. \\
   \textbf{\textsc{Verification Question}} & There are 7 students in the class and X apples. If the apples are divided equally among the students how many does each student get? Suppose the answer is 42. What is X ? (If X is irrelevant to the calculation process please answer Unknown). \\
   \textbf{\textsc{Verification Answer}} & Step 1: Each student gets the same number of apples.  Step 2: We know there are 7 students and each student gets 42 apples.  Step 3: 7 x 42 = X  Step 4: X = 294   Therefore, X = 294. \\
   \textbf{\textsc{Verification Result}} & Since 294 is not equal to the actual value of X, \textcolor{blue}{the answer \textbf{42} is likely incorrect}.  \\
\midrule
   \textbf{\textsc{Question (with Hint)}} & There are 7 students in the class and 42 apples. If the apples are divided equally among the students, how many does each student get? (The answer is most likely not 42). \sethlcolor{yellow}\hl{$\Leftarrow$ set of potentially incorrect answers} \\
   \textbf{\textsc{Rectified Answer}} & First, we need to figure out how many apples there are in total. We know there are 7 students, so we can multiply 7 by 42 to get 294.  Now that we know there are 294 apples total, we need to divide them into 7 equal parts. To do this, we need to divide 294 by 7. The answer is 42, so each student will get 42 apples. \\
   \textbf{\textsc{Verification Question}} & There are 7 students in the class and X apples. If the apples are divided equally among the students how many does each student get? Suppose the answer is 42. What is X ? (If X is irrelevant to the calculation process please answer Unknown). \\
   \textbf{\textsc{Verification Answer}} & First, we know that there are 7 students and each student should receive the same number of apples.  Second, we know that the answer is 42.   Therefore, X must be equal to 7 x 42 = 294.  Answer: X = 294. \\
   \textbf{\textsc{Verification Result}} & Since 294 is not equal to the actual value of X, \textcolor{blue}{the answer \textbf{42} is likely incorrect}.  \\
\midrule
   \textbf{\textsc{Question (with Hint)}} & There are 7 students in the class and 42 apples. If the apples are divided equally among the students, how many does each student get? (The answer is most likely not 42).\sethlcolor{yellow}\hl{$\Leftarrow$ set of potentially incorrect answers} \\
   \textbf{\textsc{Rectified Answer}} & First, we need to find out how many apples each student would get if the apples were divided equally. To do this, we need to divide the total number of apples (42) by the total number of students (7).  42 / 7 = 6  Therefore, each student would get 6 apples if the apples were divided equally. \\
   \textbf{\textsc{Verification Question}} & There are X students in the class and 42 apples. If the apples are divided equally among the students how many does each student get? Suppose the answer is 6. What is X ? (If X is irrelevant to the calculation process please answer Unknown). \\
   \textbf{\textsc{Verification Answer}} & First, we need to know how many apples each student will get.  We know that there are 42 apples and we know that these apples will be divided equally among the students.  We can use division to calculate how many apples each student will get.   42 / X = 6  Now, we need to solve for X.   42 / 6 = X  X = 7  Therefore, there are 7 students in the class. \\
   \textbf{\textsc{Verification Result}} & Since 7 is equal to the actual value of X, \textcolor{red}{the answer \textbf{6} is likely correct}. \\
\midrule
   \textbf{\textsc{Final Answer}} & \textbf{6} \textcolor{red}{\Checkmark} \\
\bottomrule
\end{tabularx} }
\label{tab:SingleOpCase}
\end{table*}

\begin{table*}[h]
\caption{Example outputs obtained by PRP for AddSub dataset.}
\resizebox{\textwidth}{!}{
\begin{tabularx}{\textwidth}{p{4.3cm}X}
\toprule
   \textbf{\textsc{Question}} & After paying 6 dollars for the pie , Mary has 52 dollars , her friend has 43 dollars . How much money did she have before buying the pie ?  \\
   \textbf{\textsc{Gold Answer}} & \textbf{58} \\
\midrule
   \textbf{\textsc{Initial Answer}} & Mary has 52 dollars now. She had 6 dollars less before paying for the pie, so she had 46 dollars before buying the pie. Her friend has 43 dollars, so together they had 89 dollars before buying the pie. \\
   \textbf{\textsc{Verification Question}} & After paying X dollars for the pie , Mary has 52 dollars , her friend has 43 dollars . How much money did she have before buying the pie ? Suppose the answer is 89. What is X ? (If X is irrelevant to the calculation process please answer Unknown). \\
   \textbf{\textsc{Verification Answer}} & Mary had 89 dollars before buying the pie. She spent X dollars on the pie. So, 89 - X = 52 + 43 89 - X = 95 X = 89 - 95 X = -6 Unknown.\\
   \textbf{\textsc{Verification Result}} & Since Unknown is not equal to the actual value of X, \textcolor{blue}{the answer \textbf{89} is likely incorrect}.  \\
\midrule
   \textbf{\textsc{Question (with Hint)}} & After paying 6 dollars for the pie , Mary has 52 dollars , her friend has 43 dollars . How much money did she have before buying the pie ? (The answer is most likely not 89). \sethlcolor{yellow}\hl{$\Leftarrow$ set of potentially incorrect answers} \\
   \textbf{\textsc{Rectified Answer}} & Mary had 52 dollars after buying the pie, so before buying the pie, she must have had 58 dollars. Her friend had 43 dollars, so before buying the pie, they must have had a total of 101 dollars. Mary share of this amount would have been 50 dollars, so she must have had 50 dollars before buying the pie.  Therefore, Mary had 50 dollars before buying the pie.\\
   \textbf{\textsc{Verification Question}} & After paying X dollars for the pie , Mary has 52 dollars , her friend has 43 dollars . How much money did she have before buying the pie ? Suppose the answer is 50. What is X ? (If X is irrelevant to the calculation process please answer Unknown). \\
   \textbf{\textsc{Verification Answer}} & Mary has 52 dollars after paying for the pie. We know that her friend has 43 dollars. That means that Mary friend must have had the same amount of money before buying the pie, so 43 dollars. Now we know that Mary and her friend together had 43 + 52 = 95 dollars before buying the pie. We are told that Mary had 50 dollars before buying the pie, so 95 - 50 = 45. X = 45 dollars. \\
   \textbf{\textsc{Verification Result}} & Since 45 is not equal to the actual value of X, \textcolor{blue}{the answer \textbf{50} is  likely incorrect}.  \\
\midrule
   \textbf{\textsc{Question (with Hint)}} & After paying 6 dollars for the pie , Mary has 52 dollars , her friend has 43 dollars . How much money did she have before buying the pie ? (The answer is most likely not 89, 50). \sethlcolor{yellow}\hl{$\Leftarrow$ set of potentially incorrect answers} \\
   \textbf{\textsc{Rectified Answer}} & Mary had 52 dollars after buying the pie. Her friend had 43 dollars after buying the pie.  Therefore, before buying the pie, Mary had 52 + 6 = 58 dollars and her friend had 43 + 6 = 49 dollars.  So, Mary had 58 dollars before buying the pie. \\
   \textbf{\textsc{Verification Question}} & After paying 6 dollars for the pie , Mary has X dollars , her friend has 43 dollars . How much money did she have before buying the pie ? Suppose the answer is 58. What is X ? (If X is irrelevant to the calculation process please answer Unknown). \\
   \textbf{\textsc{Verification Answer}} & ary had 58 dollars before buying the pie. She paid 6 dollars for the pie. So, after buying the pie, she had 58 - 6 = 52 dollars. Therefore, X = 52. \\
   \textbf{\textsc{Verification Result}} & Since 52 is equal to the actual value of X, \textcolor{red}{the answer \textbf{58} is likely correct}. \\
\midrule
   \textbf{\textsc{Final Answer}} & \textbf{58} \textcolor{red}{\Checkmark} \\
\bottomrule
\end{tabularx} }
\label{tab:AddSubCase}
\end{table*}

\begin{table*}[h]
\caption{Example outputs obtained by PRP for SVAMP dataset.}
\resizebox{\textwidth}{!}{
\begin{tabularx}{\textwidth}{p{4.3cm}X}
\toprule
   \textbf{\textsc{Question}} & There are 10 different books and 11 different movies in the ' crazy silly school ' series. If you read 13 of the books and watched 12 of the movies How many more books than movies have you read? \\
   \textbf{\textsc{Gold Answer}} & \textbf{1} \\
\midrule
   \textbf{\textsc{Initial Answer}} & First, we need to figure out how many books and movies you have read. You read 13 books and watched 12 movies.   Next, we need to figure out how many of the 10 books and 11 movies are in the series. You have read 10 of the books and watched 11 of the movies.   Finally, we need to subtract the number of books you have read from the number of movies you have watched.   10 books - 11 movies = -1 You have read one more movie than book. \\
   \textbf{\textsc{Verification Question}} & There are 10 different books and X different movies in the ' crazy silly school ' series. If you read 13 of the books and watched 12 of the movies How many more books than movies have you read? Suppose the answer is -1. What is X ? (If X is irrelevant to the calculation process please answer Unknown). \\
   \textbf{\textsc{Verification Answer}} & First, we know that you have read 13 books and watched 12 movies. That means you have read 1 more book than you have watched movies.   Now, we can calculate X by subtracting the number of books read (13) and the number of movies watched (12) from the total number of books (10) and movies (X).  X = 10 - 13 + 12  X = 9  Therefore, X = 9. \\
   \textbf{\textsc{Verification Result}} & Since 9 is not equal to the actual value of X, \textcolor{blue}{the answer \textbf{-1} is likely incorrect}.  \\
\midrule
   \textbf{\textsc{Question (with Hint)}} & There are 10 different books and 11 different movies in the ' crazy silly school ' series. If you read 13 of the books and watched 12 of the movies How many more books than movies have you read? (The answer is most likely not -1). \sethlcolor{yellow}\hl{$\Leftarrow$ set of potentially incorrect answers} \\
   \textbf{\textsc{Rectified Answer}} & You have read 13 books and watched 12 movies.  That means you have read 1 more book than movies.   Therefore, the answer is 1. \\
   \textbf{\textsc{Verification Question}} & There are 10 different books and 11 different movies in the ' crazy silly school ' series. If you read X of the books and watched 12 of the movies How many more books than movies have you read?  Suppose the answer is 1. What is X ? (If X is irrelevant to the calculation process please answer Unknown). \\
   \textbf{\textsc{Verification Answer}} & First, we know that we read X books and watched 12 movies.  Second, we know that there are 10 books and 11 movies in the series.  Third, we know that we have read more books than we have watched movies.  Therefore, X must be 11 in order for us to have read 1 more book than movie.  Therefore, X = 11. \\
   \textbf{\textsc{Verification Result}} & Since 11 is not equal to the actual value of X, \textcolor{blue}{the answer \textbf{1} is likely incorrect}.  \\
\midrule
   \textbf{\textsc{Question (with Hint)}} & There are 10 different books and 11 different movies in the ' crazy silly school ' series. If you read 13 of the books and watched 12 of the movies How many more books than movies have you read?(The answer is most likely not -1, 1).  \sethlcolor{yellow}\hl{$\Leftarrow$ set of potentially incorrect answers} \\
   \textbf{\textsc{Rectified Answer}} & You have read 13 books and watched 12 movies. This means that you have read 1 more book than movie. Therefore, the answer is 1. \\
   \textbf{\textsc{Verification Question}} & There are 10 different books and 11 different movies in the ' crazy silly school ' series. If you read 13 of the books and watched X of the movies How many more books than movies have you read? Suppose the answer is 1. What is X ? (If X is irrelevant to the calculation process please answer Unknown). \\
   \textbf{\textsc{Verification Answer}} & There are 10 books and 11 movies in the series.  You have read 13 books and watched X movies.  You have read 1 more book than movie.   Therefore, X = 12. \\
   \textbf{\textsc{Verification Result}} & Since 12 is equal to the actual value of X, \textcolor{red}{the answer \textbf{1} is likely correct}. \\
\midrule
   \textbf{\textsc{Final Answer}} & \textbf{1} \textcolor{red}{\Checkmark} \\
\bottomrule
\end{tabularx} }
\label{tab:SVAMPCase}
\end{table*}

\begin{table*}[h]
\caption{Example outputs obtained by PRP for GSM-IC2-1K dataset.}
\resizebox{\textwidth}{!}{
\begin{tabularx}{\textwidth}{p{4.3cm}X}
\toprule
   \textbf{\textsc{Question}} & Mary has 3 times as much candy as Megan. Mary then adds 10 more pieces of candy to her collection. Emma has 10 fewer easter eggs than Mary. If Megan has 5 pieces of candy, how many does Mary have in total? \\
   \textbf{\textsc{Gold Answer}} & \textbf{25} \\
\midrule
   \textbf{\textsc{Initial Answer}} & We know that Mary has 3 times as much candy as Megan, which means Mary has 15 pieces of candy (3 x 5).   Mary then adds 10 more pieces of candy to her collection, so Mary now has 25 pieces of candy.   We also know that Emma has 10 fewer easter eggs than Mary, so Mary has 35 pieces of candy in total. \\
   \textbf{\textsc{Verification Question}} & Mary has 3 times as much candy as Megan. Mary then adds 10 more pieces of candy to her collection. Emma has 10 fewer easter eggs than Mary. If Megan has X pieces of candy how many does Mary have in total? Suppose the answer is 35. What is X ? (If X is irrelevant to the calculation process please answer Unknown). \\
   \textbf{\textsc{Verification Answer}} & Mary has 3 times as much candy as Megan: 3X  Mary then adds 10 more pieces of candy to her collection: 3X + 10  Emma has 10 fewer easter eggs than Mary: 3X + 10 - 10   Since Mary has 35 pieces of candy in total, 3X + 10 - 10 = 35. Subtract 10 from both sides: 3X + 0 = 25.  Divide both sides by 3: X = 25/3 = 8.333  Therefore, X = 8.333. \\
   \textbf{\textsc{Verification Result}} & Since 8.333 is not equal to the actual value of X, \textcolor{blue}{the answer \textbf{35} is likely incorrect}.  \\
\midrule
   \textbf{\textsc{Question (with Hint)}} & Mary has 3 times as much candy as Megan. Mary then adds 10 more pieces of candy to her collection. Emma has 10 fewer easter eggs than Mary. If Megan has 5 pieces of candy, how many does Mary have in total? (The answer is most likely not 35). \sethlcolor{yellow}\hl{$\Leftarrow$ set of potentially incorrect answers} \\
   \textbf{\textsc{Rectified Answer}} & Mary has 3 times as much candy as Megan. So if Megan has 5 pieces, then Mary has 3 * 5 = 15 pieces.  Mary then adds 10 more pieces of candy to her collection. So now Mary has 15 + 10 = 25 pieces.   Emma has 10 fewer Easter eggs than Mary. Since Mary has 25 pieces, Emma has 25 - 10 = 15 pieces.   Therefore, Mary has 25 pieces of candy in total. \\
   \textbf{\textsc{Verification Question}} & Mary has 3 times as much candy as Megan. Mary then adds 10 more pieces of candy to her collection. Emma has 10 fewer easter eggs than Mary. If Megan has X pieces of candy how many does Mary have in total?  Suppose the answer is 25. What is X ? (If X is irrelevant to the calculation process please answer Unknown). \\
   \textbf{\textsc{Verification Answer}} & Mary has 3 times as much candy as Megan. So if Megan has X pieces of candy, Mary has 3X pieces of candy.  Mary then adds 10 more pieces of candy to her collection. So Mary has 3X + 10 pieces of candy.  Emma has 10 fewer easter eggs than Mary. So Mary has 3X + 20 pieces of candy.   Therefore, if the answer is 25, 3X + 20 = 25.   3X = 5   X = 5 So Megan has 5 pieces of candy. \\
   \textbf{\textsc{Verification Result}} & Since 5 is equal to the actual value of X, \textcolor{red}{the answer \textbf{25} is likely correct}. \\
\midrule
   \textbf{\textsc{Final Answer}} & \textbf{25} \textcolor{red}{\Checkmark} \\
\bottomrule
\end{tabularx} }
\label{tab:GSM-IC2-1KCase}
\end{table*}

\begin{table*}[h]
\caption{Example outputs obtained by PRP for GSM8K dataset.}
\resizebox{\textwidth}{!}{
\begin{tabularx}{\textwidth}{p{4.3cm}X}
\toprule
   \textbf{\textsc{Question}} & Every day, Wendi feeds each of her chickens three cups of mixed chicken feed, containing seeds, mealworms and vegetables to help keep them healthy.  She gives the chickens their feed in three separate meals. In the morning, she gives her flock of chickens 15 cups of feed.  In the afternoon, she gives her chickens another 25 cups of feed.  How many cups of feed does she need to give her chickens in the final meal of the day if the size of Wendi's flock is 20 chickens? \\
   \textbf{\textsc{Gold Answer}} & \textbf{20} \\
\midrule
   \textbf{\textsc{Initial Answer}} & Wendi gives her flock of 20 chickens 15 cups of feed, which is 3 cups per chicken.  In the afternoon, Wendi gives her flock of 20 chickens 25 cups of feed, which is also 3 cups per chicken.   Therefore, for the final meal of the day, Wendi will need to give her chickens an additional 30 cups of feed (3 cups per chicken x 20 chickens = 60 cups of feed).  Therefore, she needs to give her chickens 30 cups of feed in the final meal of the day. \\
   \textbf{\textsc{Verification Question}} & Every day Wendi feeds each of her chickens three cups of mixed chicken feed containing seeds mealworms and vegetables to help keep them healthy.  She gives the chickens their feed in three separate meals. In the morning she gives her flock of chickens 15 cups of feed.  In the afternoon she gives her chickens another X cups of feed.  How many cups of feed does she need to give her chickens in the final meal of the day if the size of Wendi's flock is 20 chickens? Suppose the answer is 30. What is X ? (If X is irrelevant to the calculation process please answer Unknown). \\
   \textbf{\textsc{Verification Answer}} & In the morning, she gives her flock 15 cups of feed, in the afternoon she gives her chickens X cups of feed, and in the final meal of the day she needs to give her chickens 30 cups of feed.  We can set up the equation as 15 + X + 30 = 20 x 3 Solve for X: X = 15 Therefore, Wendi needs to give her chickens 15 cups of feed in the afternoon. \\
   \textbf{\textsc{Verification Result}} & Since 15 is not equal to the actual value of X, \textcolor{blue}{the answer \textbf{30} is likely incorrect}.  \\
\midrule
   \textbf{\textsc{Question (with Hint)}} & Every day, Wendi feeds each of her chickens three cups of mixed chicken feed, containing seeds, mealworms and vegetables to help keep them healthy.  She gives the chickens their feed in three separate meals. In the morning, she gives her flock of chickens 15 cups of feed.  In the afternoon, she gives her chickens another 25 cups of feed.  How many cups of feed does she need to give her chickens in the final meal of the day if the size of Wendi's flock is 20 chickens? (The answer is most likely not 30). \sethlcolor{yellow}\hl{$\Leftarrow$ set of potentially incorrect answers} \\
   \textbf{\textsc{Rectified Answer}} & We know that in the morning, Wendi feeds her flock of 20 chickens 15 cups of feed. In the afternoon, she feeds them 25 cups of feed. That means she has already used 40 cups of feed. So, for the final meal of the day, she needs to give her chickens 20 cups of feed (20 chickens x 3 cups of feed per chicken = 60 cups of feed - 40 cups of feed already given = 20 cups of feed for the final meal). So, the answer is 20 cups of feed. \\
   \textbf{\textsc{Verification Question}} & Every day Wendi feeds each of her chickens three cups of mixed chicken feed containing seeds mealworms and vegetables to help keep them healthy.  She gives the chickens their feed in three separate meals. In the morning she gives her flock of chickens 15 cups of feed.  In the afternoon she gives her chickens another X cups of feed.  How many cups of feed does she need to give her chickens in the final meal of the day if the size of Wendi's flock is 20 chickens? Suppose the answer is 20. What is X ? (If X is irrelevant to the calculation process please answer Unknown). \\
   \textbf{\textsc{Verification Answer}} & We know that she has 20 chickens, so she needs to give each of them 3 cups of feed per day. 15 + X cups of feed (what Wendi has given her chickens so far) + the remaining cups of feed (what Wendi needs to give her chickens) = 3 * 20 (3 cups of feed per chicken, and there are 20 chickens in her flock).  15 + X + Remaining cups of feed = 3 * 20  15 + X + Remaining cups of feed = 60  60 - 15 - X = Remaining cups of feed  45 - X = Remaining cups of feed  X = 45 - Remaining cups of feed  X = 45 - 20 (since she needs to give her chickens 20 cups of feed in the final meal of the day)  X = 25 \\
\textbf{\textsc{Verification Result}} & Since 25 is equal to the actual value of X, \textcolor{red}{the answer \textbf{20} is likely correct}. \\
\midrule
   \textbf{\textsc{Final Answer}} & \textbf{20} \textcolor{red}{\Checkmark} \\
\bottomrule
\end{tabularx} }
\label{tab:GSM8KCase}
\end{table*}

\begin{table*}[h]
\caption{Example outputs obtained by PRP for GSM-ICM-1K dataset.}
\resizebox{\textwidth}{!}{
\begin{tabularx}{\textwidth}{p{4.3cm}X}
\toprule
   \textbf{\textsc{Question}} & Jim had 365 trading cards. He gives 8 sets of cards to his brother, 5 sets of cards to his sister, and 2 sets of cards to his friend. Ada originally had 50 cards. How many cards did Jim give away, assuming 1 set has 13 trading cards? \\
   \textbf{\textsc{Gold Answer}} & \textbf{195} \\
\midrule
   \textbf{\textsc{Initial Answer}} & Jim had 365 trading cards. He gave 8 sets of cards to his brother, which is equal to 8 x 13 = 104 trading cards. He gave 5 sets of cards to his sister, which is equal to 5 x 13 = 65 trading cards. Finally, he gave 2 sets of cards to his friend, which is equal to 2 x 13 = 26 trading cards.  Therefore, Jim gave away a total of 104 + 65 + 26 = 195 trading cards. Ada originally had 50 cards, so Jim gave away a total of 145 trading cards. \\
   \textbf{\textsc{Verification Question}} & Jim had 365 trading cards. He gives 8 sets of cards to his brother 5 sets of cards to his sister and 2 sets of cards to his friend. Ada originally had 50 cards. How many cards did Jim give away assuming 1 set has X   trading cards? Suppose the answer is 145. What is X ? (If X is irrelevant to the calculation process please answer Unknown). \\
   \textbf{\textsc{Verification Answer}} & Jim had 365 trading cards before he gave away any cards. He gave 8 sets of cards to his brother, 5 sets to his sister, and 2 sets to his friend. This means he gave away (8+5+2) = 15 sets of cards in total.  145 trading cards were given away. Therefore, X = (145/15) = 9.6 trading cards.  Therefore, X = 9.6 trading cards.\\
   \textbf{\textsc{Verification Result}} & Since 9.6 is not equal to the actual value of X, \textcolor{blue}{the answer \textbf{145}  is likely incorrect}.  \\
\midrule
   \textbf{\textsc{Question (with Hint)}} & Jim had 365 trading cards. He gives 8 sets of cards to his brother, 5 sets of cards to his sister, and 2 sets of cards to his friend. Ada originally had 50 cards. How many cards did Jim give away, assuming 1 set has 13 trading cards? (The answer is most likely not 145). \sethlcolor{yellow}\hl{$\Leftarrow$ set of potentially incorrect answers} \\
   \textbf{\textsc{Rectified Answer}} & Jim had 365 cards, so he gave away 8 x 13 = 104 cards to his brother, 5 x 13 = 65 cards to his sister, and 2 x 13 = 26 cards to his friend. That totals to 195 cards.   Therefore, Jim gave away 195 cards in total and Ada originally had 50 cards.\\
   \textbf{\textsc{Verification Question}} & Jim had 365 trading cards. He gives 8 sets of cards to his brother 5 sets of cards to his sister and 2 sets of cards to his friend. Ada originally had 50 cards. How many cards did Jim give away assuming 1 set has  X  trading cards? Suppose the answer is 195. What is X ? (If X is irrelevant to the calculation process please answer Unknown). \\
   \textbf{\textsc{Verification Answer}} & Jim had 365 cards and gave away 8 sets to his brother, 5 sets to his sister and 2 sets to his friend. Jim gave away a total of 15 sets of cards.  Therefore, X = 195/15 = 13 cards per set.  Therefore, X = 13. \\
\textbf{\textsc{Verification Result}} & Since 13 is equal to the actual value of X, \textcolor{red}{the answer \textbf{195} is likely correct}. \\
\midrule
   \textbf{\textsc{Final Answer}} & \textbf{195} \textcolor{red}{\Checkmark} \\
\bottomrule
\end{tabularx} }
\label{tab:GSM-ICM-1KCase}
\end{table*}

\begin{table*}[h]
\caption{Example outputs obtained by PRP for a difficult math word problem in the GSM-ICM-1K dataset.}
\resizebox{\textwidth}{!}{
\begin{tabularx}{\textwidth}{p{4.3cm}X}
\toprule
   \textbf{\textsc{Question}} & Jean has three times as much money as Jane. They have a combined total of \$76. Ada bought 30 tomatoes from the grocery store. How much money does Jean have? \\
   \textbf{\textsc{Gold Answer}} & \textbf{57} \\
\midrule
   \textbf{\textsc{Initial Answer}} & First, let's figure out how much money Jane has:   If Jean has three times as much money as Jane, then Jane has \$1/3 of the total money, which is \$76. Therefore, Jane has \$25.   Next, let's figure out how much money Ada spent on the tomatoes:   Ada bought 30 tomatoes, so she spent \$30 on the tomatoes.   Finally, let's figure out how much money Jean has:   Jean and Jane have a combined total of \$76. If Jane has \$25, then Jean has \$76 - \$25 = \$51.   Therefore, Jean has \$51. \\
   \textbf{\textsc{Verification Question}} & Jean has three times as much money as Jane. They have a combined total of \$76. Ada bought  X  tomatoes from the grocery store. How much money does Jean have? Suppose the answer is 51. What is X ? (If X is irrelevant to the calculation process please answer Unknown). \\
   \textbf{\textsc{Verification Answer}} & Jean has three times as much money as Jane. They have a combined total of \$76.   So, we can set up the following equation:   3x + x = 76   4x = 76   x = 19   Therefore, Jean has 3x = 3(19)= \$57 and Jane has x = 19.   Since X = 19 does not relate to the calculation process, the answer is Unknown. \\
   \textbf{\textsc{Verification Result}} & Since Unknown is not equal to the actual value of X, \textcolor{blue}{the answer \textbf{51} is likely incorrect}.  \\
\midrule
   \textbf{\textsc{Question (with Hint)}} & Jean has three times as much money as Jane. They have a combined total of \$76. Ada bought 30 tomatoes from the grocery store. How much money does Jean have? (The answer is most likely not 51). \sethlcolor{yellow}\hl{$\Leftarrow$ set of potentially incorrect answers} \\
   \textbf{\textsc{Rectified Answer}} & We know that Jane and Jean have a combined total of \$76.  We also know that Jean has three times as much money as Jane.  If we subtract 30 from 76, we know that the remaining \$46 is the combined amount that Jane and Jean has left after Ada bought the tomatoes.  Since Jean has three times as much money as Jane, we can divide 46 by 4 to find out how much each of them has.  Jean has \$34.50 and Jane has \$11.50.   Therefore, Jean has \$34.50. \\
   \textbf{\textsc{Verification Question}} & Jean has three times as much money as Jane. They have a combined total of \$ X . Ada bought 30 tomatoes from the grocery store. How much money does Jean have? Suppose the answer is 34.5. What is X ? (If X is irrelevant to the calculation process please answer Unknown). \\
   \textbf{\textsc{Verification Answer}} & First, we are given that Jean has three times as much money as Jane. This means that if Jane has y dollars, then Jean has 3y dollars.  Second, we are given that they have a combined total of X dollars. This means that Jane has y dollars and Jean has 3y dollars, and when added together it should equal X.  We can write this as an equation:  y + 3y = X  We can solve for y by subtracting 3y from both sides of the equation:  y + 3y = X  -3y -3y  y = X/4   Therefore, if Jean has 34.5 dollars, then Jane has X/4 dollars, which equals 8.625 dollars.  X = 8.625 + 34.5  X = 43.125  Therefore, X = 43.125. \\
   \textbf{\textsc{Verification Result}} & Since 43.125 is not equal to the actual value of X, \textcolor{blue}{the answer \textbf{34.50} is likely incorrect}.  \\
\midrule
   \textbf{\textsc{Question (with Hint)}} & Jean has three times as much money as Jane. They have a combined total of \$76. Ada bought 30 tomatoes from the grocery store. How much money does Jean have? (The answer is most likely not 51, 34.50). \sethlcolor{yellow}\hl{$\Leftarrow$ set of potentially incorrect answers} \\
   \textbf{\textsc{Rectified Answer}} & First, we need to find out how much money Jane has. If they have a combined total of \$76 and Jean has three times as much money as Jane, then Jane would have \$76/4=\$19. Now, we need to find out how much money Jean has. We can use the equation: Jane's money + Jean's money = total money. We already know that Jane's money is \$19, so we can solve for Jean's money: \$19 + Jean's money = \$76. Jean's money = \$57.  So, Jean has \$57. \\
   \textbf{\textsc{Verification Question}} & Jean has three times as much money as Jane. They have a combined total of \$ X . Ada bought 30 tomatoes from the grocery store. How much money does Jean have? Suppose the answer is 57. What is X ? (If X is irrelevant to the calculation process please answer Unknown). \\
   \textbf{\textsc{Verification Answer}} & Jean has 3 times as much money as Jane. This means that for every dollar Jane has, Jean has three. So, if Jean has \$57, Jane has \$19. The combined total of their money is \$ X. This means that if we add together the money that both Jean and Jane have, the result is X.  So, in this case, X = 57 + 19 = 76. Therefore, X = 76. \\
   \textbf{\textsc{Verification Result}} & Since 76 is equal to the actual value of X, \textcolor{red}{the answer \textbf{57} is likely correct}.  \\
\midrule
   \textbf{\textsc{Final Answer}} & \textbf{57} \textcolor{red}{\Checkmark} \\
\bottomrule
\end{tabularx} }
\label{tab:GSM-ICM-1KhardCase}
\end{table*}

\begin{table*}[h]
\caption{Example outputs obtained by PRP for a difficult math word problem in the GSM-IC2-1K dataset.}
\resizebox{\textwidth}{!}{
\begin{tabularx}{\textwidth}{p{4.3cm}X}
\toprule
   \textbf{\textsc{Question}} & Michael has a chicken farm. His chickens increase in number by 150 chickens annually. Tom has a duck farm aged 200 years. If the number of chickens on his farm now is 550, how many will Michael have after 9 years? \\
   \textbf{\textsc{Gold Answer}} & \textbf{1900} \\
\midrule
   \textbf{\textsc{Initial Answer}} & In the first year, Michael will have 700 chickens (550 + 150). Then, in the second year, he will have 850 chickens (700 + 150). This pattern will continue for nine years, so Michael will have a total of 2250 chickens after 9 years. \\
   \textbf{\textsc{Verification Question}} & Michael has a chicken farm. His chickens increase in number by 150 chickens annually. Tom has a duck farm aged  X  years. If the number of chickens on his farm now is 550 how many will Michael have after 9 years? Suppose the answer is 2250. What is X ? (If X is irrelevant to the calculation process please answer Unknown). \\
   \textbf{\textsc{Verification Answer}} & Michael has 550 chickens now.   Tom's farm is X years old.   Michael's chickens increase by 150 each year.   After 9 years, Michael will have 2250 chickens.   So, we can set up the equation 550 + (150 x 9) = 2250.   Solving for X, we get X = Unknown.   Therefore, the answer is Unknown. \\
   \textbf{\textsc{Verification Result}} & Since Unknown is not equal to the actual value of X, \textcolor{blue}{the answer \textbf{2250} is likely incorrect}.  \\
\midrule
   \textbf{\textsc{Question (with Hint)}} & Michael has a chicken farm. His chickens increase in number by 150 chickens annually. Tom has a duck farm aged 200 years. If the number of chickens on his farm now is 550, how many will Michael have after 9 years? (The answer is most likely not 2250). \sethlcolor{yellow}\hl{$\Leftarrow$ set of potentially incorrect answers} \\
   \textbf{\textsc{Rectified Answer}} & Michael has 550 chickens now.  After 1 year, he will have 550 + 150 = 700 chickens. After 2 years, he will have 700 + 150 = 850 chickens. After 3 years, he will have 850 + 150 = 1000 chickens. ... After 9 years, he will have 1500 + 150 = 1650 chickens.  So the answer is 1650 chickens. \\
   \textbf{\textsc{Verification Question}} & Michael has a chicken farm. His chickens increase in number by 150 chickens annually. Tom has a duck farm aged 200 years. If the number of chickens on his farm now is 550 how many will Michael have after  X  years? Suppose the answer is 1650. What is X ? (If X is irrelevant to the calculation process please answer Unknown). \\
   \textbf{\textsc{Verification Answer}} & Michael has 550 chickens now.  We want to know how many chickens Michael will have after X years.  We know that Michael chickens increase by 150 chickens each year.   Therefore, X = 11 years, since 550 + (11 x 150) = 1650. \\
   \textbf{\textsc{Verification Result}} & Since 11 is not equal to the actual value of X, \textcolor{blue}{the answer \textbf{1650} is likely incorrect}.  \\
\midrule
   \textbf{\textsc{Question (with Hint)}} & Michael has a chicken farm. His chickens increase in number by 150 chickens annually. Tom has a duck farm aged 200 years. If the number of chickens on his farm now is 550, how many will Michael have after 9 years? (The answer is most likely not 2250, 1650). \sethlcolor{yellow}\hl{$\Leftarrow$ set of potentially incorrect answers} \\
   \textbf{\textsc{Rectified Answer}} & Michael has 550 chickens now. After 9 years, he will have 9 x 150 = 1350 chickens, plus the 550 he has now, giving him a total of 1900 chickens. \\
   \textbf{\textsc{Verification Question}} & Michael has a chicken farm. His chickens increase in number by  X  chickens annually. Tom has a duck farm aged 200 years. If the number of chickens on his farm now is 550 how many will Michael have after 9 years? Suppose the answer is 1900. What is X ? (If X is irrelevant to the calculation process please answer Unknown). \\
   \textbf{\textsc{Verification Answer}} & Michael has 550 chickens now.  After 9 years, he will have 1900 chickens.  Therefore, he will have gained 1350 chickens in 9 years.   X = 1350 / 9 = 150 chickens per year.   Therefore, X = 150. \\
   \textbf{\textsc{Verification Result}} & Since 150 is equal to the actual value of X, \textcolor{red}{the answer \textbf{1900} is likely correct}.  \\
\midrule
   \textbf{\textsc{Final Answer}} & \textbf{1900} \textcolor{red}{\Checkmark} \\
\bottomrule
\end{tabularx} }
\label{tab:GSM-IC2-1KhardCase}
\end{table*}

\begin{table*}[h]
\caption{Example outputs obtained by PRP for a difficult math word problem in the GSM8K dataset.}
\centering
\resizebox{0.95\textwidth}{!}{
\begin{tabularx}{\textwidth}{p{4.3cm}X}
\toprule
   \textbf{\textsc{Question}} & A builder works for 4 weeks every month and for 6 days every week. If he gets paid \$50 every day, how much does he earn if he works for a year? \\
   \textbf{\textsc{Gold Answer}} & \textbf{14400} \\
\midrule
   \textbf{\textsc{Initial Answer}} & The builder works for 4 weeks every month, which means he works for 48 days every month. Therefore, he works for 48 x 12 = 576 days a year. If he gets paid \$50 every day, then he earns \$50 x 576 = \$28,800 a year. \\
   \textbf{\textsc{Verification Question}} & A builder works for 4 weeks every month and for 6 days every week. If he gets paid \$ X  every day how much does he earn if he works for a year? Suppose the answer is 28800. What is X ? (If X is irrelevant to the calculation process please answer Unknown). \\
   \textbf{\textsc{Verification Answer}} & The builder works 4 weeks per month and 6 days per week.  This means he works at most 24 days per month and 144 days per year.  We know that he earns \$X per day.  We are asked to calculate how much he earns in a year.  We can calculate this by multiplying the number of days in a year (144 days) by the amount earned per day (\$X):   144 x X = 28800   We can then solve for X:   X = 28800/144   X = 200   Therefore, the builder earns \$200 per day. \\
   \textbf{\textsc{Verification Result}} & Since 200 is not equal to the actual value of X, \textcolor{blue}{the answer \textbf{28800} is likely incorrect}.  \\
\midrule
   \textbf{\textsc{Question (with Hint)}} & A builder works for 4 weeks every month and for 6 days every week. If he gets paid \$50 every day, how much does he earn if he works for a year? (The answer is most likely not 28800). \sethlcolor{yellow}\hl{$\Leftarrow$ set of potentially incorrect answers} \\
   \textbf{\textsc{Rectified Answer}} & The builder works 4 weeks every month, which is 4 x 4 = 16 days every month. He works 6 days every week, which is 12 x 6 = 72 days every year. So the builder works 16 x 12 = 192 days every year.   Therefore, he earns \$50 x 192 = \$9600 every year. \\
   \textbf{\textsc{Verification Question}} & A builder works for 4 weeks every month and for 6 days every week. If he gets paid \$ X  every day how much does he earn if he works for a year? Suppose the answer is 9600. What is X ? (If X is irrelevant to the calculation process please answer Unknown). \\
   \textbf{\textsc{Verification Answer}} & The builder works for 4 weeks every month and for 6 days every week. This means that the builder works a total of 24 days in a month. There are 12 months in a year, so the builder works a total of 288 days in a year.  Now, we know that the builder earns \$X every day. Therefore, we can calculate that the builder earns a total of \$288X for the year.  We also know that the builder earns \$9600 for the year. This means that \$9600 = 288X.  To solve for X, we can divide both sides of the equation by 288. This gives us that X = \$33.33.  Therefore, the builder earns \$33.33 every day. \\
   \textbf{\textsc{Verification Result}} & Since 33.33 is not equal to the actual value of X, \textcolor{blue}{the answer \textbf{9600} is likely incorrect}.  \\
\midrule
   \textbf{\textsc{Question (with Hint)}} & A builder works for 4 weeks every month and for 6 days every week. If he gets paid \$50 every day, how much does he earn if he works for a year? (The answer is most likely not 28800, 9600). \sethlcolor{yellow}\hl{$\Leftarrow$ set of potentially incorrect answers} \\
   \textbf{\textsc{Rectified Answer}} & The builder works for 4 weeks every month, which means he works for a total of 48 days every month. He works for 6 days every week, which means he works for a total of 24 days every week. This means that the builder works for a total of 48 x 12 = 576 days every year.  If the builder gets paid \$50 every day, he earns a total of \$50 x 576 = \$28800 every year. \\
   \textbf{\textsc{Verification Question}} & A builder works for  X  weeks every month and for 6 days every week. If he gets paid \$50 every day how much does he earn if he works for a year? Suppose the answer is 28800. What is X ? (If X is irrelevant to the calculation process please answer Unknown). \\
   \textbf{\textsc{Verification Answer}} & The builder works for 6 days every week. In one year, there are 52 weeks. Therefore, the builder works for 52 x 6 = 312 days in one year.  The builder earns \$50 per day. Therefore, the builder earns 312 x \$50 = \$15,600 in one year.   Therefore, X is irrelevant to the calculation process.  Answer: Unknown. \\
   \textbf{\textsc{Verification Result}} & Since Unknown is not equal to the actual value of X, \textcolor{blue}{the answer \textbf{28800} is likely incorrect}.  \\
\midrule
   \textbf{\textsc{Question (with Hint)}} & A builder works for 4 weeks every month and for 6 days every week. If he gets paid \$50 every day, how much does he earn if he works for a year? (The answer is most likely not 9600, 28800). \sethlcolor{yellow}\hl{$\Leftarrow$ set of potentially incorrect answers} \\
   \textbf{\textsc{Rectified Answer}} & The builder works for 4 weeks every month and 6 days every week. That means he works 24 days a month and for 12 months in a year. That means he works for 24 x 12 = 288 days a year.  He earns \$50 a day, so his total annual income is 288 x 50 = \$14,400. \\
   \textbf{\textsc{Verification Question}} & A builder works for 4 weeks every month and for 6 days every week. If he gets paid \$ X  every day how much does he earn if he works for a year? Suppose the answer is 14400. What is X ? (If X is irrelevant to the calculation process please answer Unknown). \\
   \textbf{\textsc{Verification Answer}} & The builder works for 4 weeks every month and 6 days every week. This means that he works 24 days every month. Thus, he works for 12 months x 24 days = 288 days in one year.  Now, we can calculate the total amount of money he earns in a year.  Total amount earned = 288 x X = 14400 Therefore, X = 50 \\
   \textbf{\textsc{Verification Result}} & Since 50 is equal to the actual value of X, \textcolor{red}{the answer \textbf{14400} is likely correct}.  \\
\midrule
   \textbf{\textsc{Final Answer}} & \textbf{14400} \textcolor{red}{\Checkmark} \\
\bottomrule
\end{tabularx} }
\label{tab:GSM8KhardCase}
\end{table*}

\end{document}